\documentclass[10pt,twocolumn,letterpaper]{article}
\usepackage{iccv}
\usepackage{times}
\usepackage{epsfig}
\usepackage{graphicx}
\usepackage{amsmath}
\usepackage{amssymb}
\usepackage[breaklinks=true,bookmarks=false]{hyperref}
\usepackage{booktabs}
\usepackage{tabularx}
\usepackage{multirow}
\usepackage{bm}
\usepackage[dvipsnames,svgnames,x11names]{xcolor}
\usepackage{enumitem}
\usepackage{soul}

\newcommand{\proposed}{\texttt{FUNIT}\xspace}
\newcommand{\scaption}[1]{\caption{#1}}
\newcommand{\mysubsection}[1]{\vspace{0.5mm}\noindent{\bf #1}}
\newcolumntype{L}[1]{>{\raggedright\arraybackslash}p{#1}}
\newcolumntype{C}[1]{>{\centering\arraybackslash}p{#1}}
\newcolumntype{R}[1]{>{\raggedleft\arraybackslash}p{#1}}
\iccvfinalcopy 
\def\iccvPaperID{3116} 

\ificcvfinal\fi
\begin{document}
\title{Few-Shot Unsupervised Image-to-Image Translation}
\author{Ming-Yu Liu$^{1}$, Xun Huang$^{1,2}$, Arun Mallya$^{1}$, Tero Karras$^{1}$, Timo Aila$^{1}$, Jaakko Lehtinen$^{1,3}$, Jan Kautz$^{1}$\\
	$^{1}$NVIDIA, $^{2}$Cornell University, $^{3}$Aalto University\\
	{\tt\small \{mingyul, xunh, amallya, tkarras, taila, jlehtinen, jkautz\}@nvidia.com}
}
\maketitle
\begin{abstract}
Unsupervised image-to-image translation methods learn to map images in a given class to an analogous image in a different class, drawing on unstructured (non-registered) datasets of images. While remarkably successful, current methods require access to many images in both source and destination classes at training time. We argue this greatly limits their use. Drawing inspiration from the human capability of picking up the essence of a novel object from a small number of examples and generalizing from there, we seek a few-shot, unsupervised image-to-image translation algorithm that works on previously unseen target classes that are specified, at test time, only by a few example images. Our model achieves this few-shot generation capability by coupling an adversarial training scheme with a novel network design. Through extensive experimental validation and comparisons to several baseline methods on benchmark datasets, we verify the effectiveness of the proposed framework. Our implementation and datasets are available at \url{https://github.com/NVlabs/FUNIT}.
\end{abstract}

\vspace{-3mm}

\section{Introduction}

Humans are remarkably good at generalization. When given a picture of a previously unseen exotic animal, say, we can form a vivid mental picture of the same animal in a different pose, especially when we have encountered (images of) similar but different animals in that pose before. For example, a person seeing a standing tiger for the first time will have no trouble imagining what it will look lying down, given a lifetime of experience of other animals. 

While recent unsupervised image-to-image translation algorithms are remarkably successful in transferring complex appearance changes across image classes \cite{liu2016coupled,taigman2017unsupervised,liu2017unsupervised,kim2017learning,zhu2017unpaired,yi2017dualgan}, the capability to generalize from few samples of a new class based on prior knowledge is entirely beyond their reach. Concretely, they need large training sets over all classes of images they are to perform translation on, i.e., they do not support few-shot generalization.

As an attempt to bridge the gap between human and machine imagination capability, we propose the Few-shot UNsupervised Image-to-image Translation (\proposed) framework, aiming at learning an image-to-image translation model for mapping an image of a source class to an analogous image of a target class by leveraging few images of the target class given at test time. The model is never shown images of the target class during training but is asked to generate some of them at test time. To proceed, we first hypothesize that the few-shot generation capability of humans develops from their past visual experiences---a person can better imagine views of a new object if the person has seen many more different object classes in the past. Based on the hypothesis, we train our \proposed model using a dataset containing images of many different object classes for simulating the past visual experiences. Specifically, we train the model to translate images from one class to another class by leveraging few example images of the another class. We hypothesize that by learning to extract appearance patterns from the few example images for the translation task, the model learns a generalizable appearance pattern extractor that can be applied to images of unseen classes at test time for the few-shot image-to-image translation task. In the experiment section, we give empirical evidence that the few-shot translation performance improves as the number of classes in the training set increases.

Our framework is based on Generative Adversarial Networks (GAN)~\cite{goodfellow2014generative}. We show that by coupling an adversarial training scheme with a novel network design we achieve the desired few-shot unsupervised image-to-image translation capability. Through extensive experimental validation on three datasets, including comparisons to several baseline methods using a variety of performance metrics, we verify the effectiveness of our proposed framework. In addition, we show the proposed framework can be applied to the few-shot image classification task. By training a classifier on the images generated by our model for the few-shot classes, we are able to outperform a state-of-the-art few-shot classification method that is based on feature hallucination. 

\vspace{-2mm}
\section{Related Work}
\vspace{-2mm}
\begin{figure*}[tbh!]
	\centering
	\includegraphics[trim=0.00in 0.0in 0in 0in, width=0.99\textwidth]{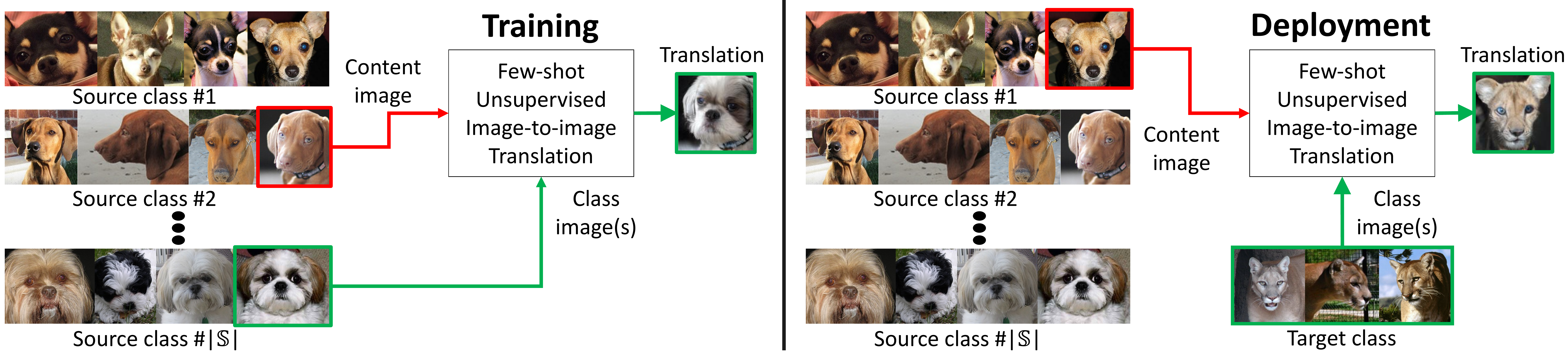}
	\vspace{2mm}
	\scaption{{\bf Training.} The training set consists of images of various object classes (source classes). We train a model to translate images between these source object classes. {\bf Deployment.} We show our trained model very few images of the target class, which is sufficient to translate images of source classes to analogous images of the target class even though the model has never seen a single image from the target class during training. Note that the \proposed generator takes two inputs: 1) a content image and 2) a set of target class images. It aims to generate a translation of the input image that resembles images of the target class.}
	\label{fig::problem_setting}
\end{figure*}

\mysubsection{Unsupervised/unpaired image-to-image translation} aims at learning a conditional image generation function that can map an input image of a source class to an analogues image of a target class without pair supervision. This problem is inherently ill-posed as it attempts to recover the joint distribution using samples from marginal distributions~\cite{liu2017unsupervised,liu2016coupled}. To deal with the problem, existing works use additional constraints. For example, some works enforce the translation to preserve certain properties of the source data, such as pixel values~\cite{shrivastava2017learning}, pixel gradients~\cite{bousmalis2017unsupervised}, semantic features~\cite{taigman2017unsupervised}, class labels \cite{bousmalis2017unsupervised}, or pairwise sample distances~\cite{benaim2017one}. There are works enforcing the cycle consistency constraint~\cite{yi2017dualgan,zhu2017unpaired,kim2017learning,almahairi2018augmented,zhu2017toward}. Several works use the shared/partially-shared latent space assumption~\cite{liu2017unsupervised,liu2016coupled}/\cite{huang2018multimodal,lee2018diverse}. Our work is based on the partially-shared latent space assumption but is designed for the few-shot unsupervised image-to-image translation task.

While capable of generating realistic translation outputs, existing unsupervised image-to-image translation models are limited in two aspects. First, they are sample inefficient, generating poor translation outputs if only few images are given at training time. Second, the learned models are limited for translating images between two classes. A trained model for one translation task cannot be directly reused for a new task despite similarity between the new task and the original task. For example, a husky-to-cat translation model can not be re-purposed for husky-to-tiger translation even though cat and tiger share a great similarity.

Recently, Benaim and Wolf~\cite{benaim2018one} proposed an unsupervised image-to-image translation framework for partially addressing the first aspect. Specifically, they use a training dataset consisting of one source class image but many target class images to train a model for translating the \textit{single} source class image to an analogous image of the target class. Our work differs from their work in several major ways. First, we assume many source class images but few target class images. Moreover, we assume that the few target class images are only available at test time and can be from many different object classes. 

\mysubsection{Multi-class unsupervised image-to-image translation}~\cite{choi2017stargan,anoosheh2017combogan,hui2017unsupervised} extends the unsupervised image-to-image translation methods to multiple classes. Our work is similar to these methods in the sense that our training dataset consists of images of multiple classes. But instead of translating images among \textit{seen} classes, we focus on translating images of seen classes to analogous images of previously \textit{unseen} classes.

\mysubsection{Few-shot classification.} Unlike few-shot image-to-image translation, the task of learning classifiers for novel classes using few examples is a long-studied problem. Early works use generative models of appearance that share priors across classes in a hierarchical manner~\cite{fei2006one,salakhutdinov2012one}. More recent works focus on using meta-learning to quickly adapt models to novel tasks~\cite{finn2017model,nichol2018first,ravi2016optimization,munkhdalai2017meta}. These methods learn better optimization strategies for training, so that the performance upon seeing only  few examples is improved. Another set of works focus on learning image embeddings that are better suited for few-shot learning~\cite{vinyals2016matching,snell2017prototypical,yang2018learning}. Several recent works propose augmenting the training set for the few-shot classification task by generating new feature vectors corresponding to novel classes~\cite{dixit2017aga,hariharan2017low,wang2018low}. Our work is designed for few-shot unsupervised image-to-image translation. However, it can be applied to few-shot classification, as shown in the experiments section.
\section{Few-shot Unsupervised Image Translation}

The proposed \proposed framework aims at mapping an image of a source class to an analogous image of an unseen target class by leveraging a few target class images that are made available at test time. To train \proposed, we use images from a set of object classes~(\eg images of various animal species), called the source classes. We do not assume existence of paired images between any two classes (\ie no two animals of different species are at exactly the same pose). We use the source class images to train a multi-class unsupervised image-to-image translation model. During testing, we provide the model few images from a novel object class, called the target class. The model has to leverage the few target images to translate any source class image to analogous images of the target class. When we provide the same model few images from a different novel object class, it has to translate any source class images to analogous images of the different novel object class. 

Our framework consists of a conditional image generator $G$ and a multi-task adversarial discriminator $D$. Unlike the conditional image generators in existing unsupervised image-to-image translation frameworks~\cite{zhu2017unpaired,liu2017unsupervised}, which take one image as input, our generator $G$ simultaneously takes \textit{a content image} $\mathbf{x}$ and \textit{a set of $K$ class images} $\{\mathbf{y}_1,...,\mathbf{y}_K\}$ as input and produce the output image $\bar{\mathbf{x}}$ via  
\begin{equation}
\bar{\mathbf{x}} = G(\mathbf{x}, \{\mathbf{y}_1,...,\mathbf{y}_K\}).
\label{eqn::output1}
\end{equation}
We assume the content image belongs to object class $c_x$ while each of the $K$ class images belong to object class $c_y$. In general, $K$ is a small number and $c_x$ is different from $c_y$. We will refer $G$ as the few-shot image translator.

As shown in Figure~\ref{fig::problem_setting}, $G$ maps an input content image $\mathbf{x}$ to an output image $\bar{\mathbf{x}}$, such that $\bar{\mathbf{x}}$ looks like an image belonging to object class $c_y$, and $\bar{\mathbf{x}}$ and $\mathbf{x}$ share structural similarity. Let $\mathbb{S}$ and $\mathbb{T}$ denote the set of source classes and the set of target classes, respectively. During training, $G$ learns to translate images between two randomly sampled source classes $c_x,c_y \in \mathbb{S}$ with $c_x \neq c_y$. At test time, $G$ takes a few images from an unseen target class $c \in \mathbb{T}$ as the class images, and maps an image sampled from any of the source classes to an analogous image of the target class $c$.

Next, we discuss the network design and learning. More details are given in Appendix~\ref{sec::arch}.

\subsection{Few-shot Image Translator}

The few-shot image translator $G$ consists of a content encoder $E_x$, a class encoder $E_y$, and a decoder $F_x$. The content encoder is made of several 2D convolutional layers followed by several residual blocks~\cite{he2016deep,johnson2016perceptual}. It maps the input content image $\mathbf{x}$ to a content latent code $\mathbf{z}_x$, which is a spatial feature map. The class encoder consists of several 2D convolutional layers followed by a mean operation along the sample axis. Specifically, it first maps each of the $K$ individual class images $\{\mathbf{y}_1,...,\mathbf{y}_K\}$ to an intermediate latent vector and then computes the mean of the intermediate latent vectors to obtain the final class latent code $\mathbf{z}_y$.

The decoder consists of several adaptive instance normalization (AdaIN) residual blocks~\cite{huang2018multimodal} followed by a couple of upscale convolutional layers. The AdaIN residual block is a residual block using the AdaIN~\cite{huang2017adain} as the normalization layer. For each sample, AdaIN first normalizes the activations of a sample in each channel to have a zero mean and unit variance. It then scales the activations using a learned affine transformation consisting of a set of scalars and biases. Note that the affine transformation is spatially invariant and hence can only be used to obtain global appearance information. The affine transformation parameters are adaptively computed using $\mathbf{z}_y$ via a two-layer fully connected network. With $E_x$, $E_y$, and $F_x$, (\ref{eqn::output1}) becomes
\begin{equation}
\bar{\mathbf{x}} = F_x\big{(}\mathbf{z}_x, \mathbf{z}_y\big{)} = F_x\big{(}E_x(\mathbf{x}), E_y(\{\mathbf{y}_1,...,\mathbf{y}_K\})\big{)}.
\end{equation}
By using this translator design, we aim at extracting class-invariant latent representation (\eg, object pose) using the content encoder and extracting class-specific latent representation (\eg, object appearance) using the class encoder. By feeding the class latent code to the decoder via the AdaIN layers, we let the class images control the global look (\eg, object appearance), while the content image determines the local structure (\eg, locations of eyes). 

At training time, the class encoder learns to extract class-specific latent representation from the images of the source classes. At test time, this generalizes to images of previously unseen classes. In the experiment section, we show that the generalization capability depends on the number of source object classes seen during training. When $G$ is trained with more source classes (\eg, more species of animals), it has a better few-shot image translation performance (\eg, better in translating husky to mountain lion).

\subsection{Multi-task Adversarial Discriminator}

Our discriminator $D$ is trained by solving multiple adversarial classification tasks simultaneously. Each of the tasks is a binary classification task determining whether an input image is a real image of the source class or a translation output coming from $G$. As there are $|\mathbb{S}|$ source classes, $D$ produces $|\mathbb{S}|$ outputs. When updating $D$ for a real image of source class $c_x$, we penalize $D$ if its $c_x$th output is false. For a translation output yielding a fake image of source class $c_x$, we penalize $D$ if its $c_x$th output is positive. We do not penalize $D$ for not predicting false for images of other classes ($\mathbb{S}\setminus\{c_x\}$). When updating $G$, we only penalize $G$ if the $c_x$th output of $D$ is false. We empirically find this discriminator works better than a discriminator trained by solving a much harder $|\mathbb{S}|$-class classification problem.

\begin{table*}[bth!]
    \centering
    \resizebox{1.00\linewidth}{!}{\mbox{%
	{\tabcolsep=10pt\def\arraystretch{1}
		\begin{tabular}{|c|c||cccc|c|cc|c|}
			\hline
			& Setting & \bf Top1-all $\uparrow$ & \bf Top5-all $\uparrow$ & \bf Top1-test $\uparrow$ & \bf Top5-test $\uparrow$ & \bf DIPD $\downarrow$& \bf IS-all $\uparrow$ & \bf IS-test $\uparrow$ & \bf mFID $\downarrow$\\
			\hline
			\parbox[t]{2mm}{\multirow{18}{*}{\rotatebox[origin=c]{90}{\textbf{\large{Animal Faces}}}}}
			& \texttt{CycleGAN-Unfair-20}&  28.97&  47.88&  38.32&  71.82&  1.615&  10.48&  7.43&  197.13\\
			& \texttt{UNIT-Unfair-20}&  22.78&  43.55&  35.73&  70.89&  1.504&  12.14&  6.86&  197.13\\
			& \texttt{MUNIT-Unfair-20}&  38.61&  62.94&  53.90&  84.00&  1.700&  10.20&  7.59&  158.93\\
			\cline{2-10}
			& \texttt{StarGAN-Unfair-1}&  2.56&  10.50&  9.07&  32.55&  1.311&  10.49&  5.17&  201.58 \\
			& \texttt{StarGAN-Unfair-5}&  12.99&  35.56&  25.40&  60.64&  1.514&  7.46&  6.10&  204.05 \\
			& \texttt{StarGAN-Unfair-10}&  20.26&  45.51&  30.26&  68.78&  1.559&  7.39&  5.83&  208.60 \\
			& \texttt{StarGAN-Unfair-15}&  20.47&  46.46&  34.90&  71.11&  1.558&  7.20&  5.58&  204.13 \\
			& \texttt{StarGAN-Unfair-20}&  24.71&  48.92&  35.23&  73.75&  1.549&  8.57&  6.21&  198.07 \\
			\cline{2-10}
			& \texttt{StarGAN-Fair-1}&  0.56&  3.46&  4.41&  20.03&  1.368&  7.83&  3.71&  228.74 \\
			& \texttt{StarGAN-Fair-5}&  0.60&  3.56&  4.38&  20.12&  1.368&  7.80&  3.72&  235.66\\
			& \texttt{StarGAN-Fair-10}&  0.60&  3.40&  4.30&  20.00&  1.368&  7.84&  3.71&  241.77\\
			& \texttt{StarGAN-Fair-15}&  0.62&  3.49&  4.28&  20.24&  1.368&  7.82&  3.72&  228.42\\
			& \texttt{StarGAN-Fair-20}&  0.62&  3.45&  4.41&  20.00&  1.368&  7.83&  3.72&  228.57\\
			\cline{2-10}
			& \texttt{FUNIT-1}&  17.07&  54.11&  46.72&  82.36&  1.364&  22.18&  10.04&  93.03 \\
			& \texttt{FUNIT-5}&  33.29&  78.19&  68.68&  96.05&  1.320&  22.56&  13.33&  70.24 \\
			& \texttt{FUNIT-10}&  37.00&  82.20&  72.18&  97.37&  1.311&  22.49&  14.12&  67.35 \\
			& \texttt{FUNIT-15}&  38.83&  83.57&  73.45&  97.77&  1.308&  22.41&  14.55&  66.58 \\
			& \texttt{FUNIT-20}&  {\bf 39.10}&  {\bf 84.39}&  {\bf 73.69}&  {\bf 97.96}&  {\bf 1.307}&  {\bf 22.54}&  {\bf 14.82}&  {\bf 66.14}\\
			\hline\hline
			\parbox[t]{2mm}{\multirow{18}{*}{\rotatebox[origin=c]{90}{\textbf{\large{North American Birds}}}}}
			& \texttt{CycleGAN-Unfair-20}&  9.24&  22.37&  19.46&  42.56&  1.488&  25.28&  7.11&  215.30\\
			& \texttt{UNIT-Unfair-20}&  7.01&  18.31&  16.66&  37.14&  1.417&  28.28&  7.57&  203.83\\
			& \texttt{MUNIT-Unfair-20}&  23.12&  41.41&  38.76&  62.71&  1.656&  24.76&  9.66&  198.55\\
			\cline{2-10}
			& \texttt{StarGAN-Unfair-1}&  0.92&  3.83&  3.98&  13.73&  1.491&  14.80&  4.10&  266.26\\
			& \texttt{StarGAN-Unfair-5}&  2.54&  8.94&  8.82&  23.98&  1.574&  13.84&  4.21&  270.12\\
			& \texttt{StarGAN-Unfair-10}&  4.26&  13.28&  12.03&  32.02&  1.571&  15.03&  4.09&  278.94\\
			& \texttt{StarGAN-Unfair-15}&  3.70&  11.74&  12.90&  31.62&  1.509&  18.61&  5.25&  252.80\\
			& \texttt{StarGAN-Unfair-20}&  5.38&  16.02&  13.95&  33.96&  1.544&  18.94&  5.24&  260.04\\
			\cline{2-10}
			& \texttt{StarGAN-Fair-1}&  0.24&  1.17&  0.97&  4.84&  1.423&  13.73&  4.83&  244.65\\
			& \texttt{StarGAN-Fair-5}&  0.22&  1.07&  1.00&  4.86&  1.423&  13.72&  4.82&  244.40\\
			& \texttt{StarGAN-Fair-10}&  0.24&  1.13&  1.03&  4.90&  1.423&  13.72&  4.83&  244.55\\
			& \texttt{StarGAN-Fair-15}&  0.23&  1.05&  1.04&  4.90&  1.423&  13.72&  4.81&  244.80\\
			& \texttt{StarGAN-Fair-20}&  0.23&  1.08&  1.00&  4.86&  1.423&  13.75&  4.82&  244.71\\
			\cline{2-10}
			& \texttt{FUNIT-1}&  11.17&  34.38&  30.86&  60.19&  1.342&  67.17&  17.16&  113.53\\
			& \texttt{FUNIT-5}&  20.24&  51.61&  45.40&  75.75&  1.296&  74.81&  22.37&  99.72\\
			& \texttt{FUNIT-10}&  22.45&  54.89&  48.24&  77.66&  1.289&  75.40&  23.60&  98.75\\
			& \texttt{FUNIT-15}&  23.18&  55.63&  49.01&  78.70&  1.287&  \textbf{76.44}&  23.86&  98.16\\
			& \texttt{FUNIT-20}&  \textbf{23.50}&  \textbf{56.37}&  \textbf{49.81}&  \textbf{78.89}&  \textbf{1.286}&  76.42& \textbf{ 24.00}&  \textbf{97.94}\\
			\hline
		\end{tabular}}%
		}}\vspace{2mm}
    	\scaption{Performance comparison with the fair and unfair baselines. $\uparrow$ means larger numbers are better, $\downarrow$ means smaller numbers are better.}\label{tbl::main_results}
	\end{table*}	

\subsection{Learning}

We train the proposed \proposed framework by solving a minimax optimization problem given by
\begin{align}
\min_{D}\max_{G} \mathcal{L}_{\text{GAN}}(D,G) + 
\lambda_{\text{R}} \mathcal{L}_{\text{R}}(G) + \lambda_{\text{F}} \mathcal{L}_{\text{FM}}(G)\label{eqn::learning}
\end{align}
where $\mathcal{L}_{\text{GAN}}$, $\mathcal{L}_{\text{R}}$, and $\mathcal{L}_{\text{F}}$ are the GAN loss, the content image reconstruction loss, and the feature matching loss. The GAN loss is a conditional one given by
\begin{align}
\mathcal{L}_{\text{GAN}}(G,D) = 
&E_{\mathbf{x}}\left[ -\log D^{c_x}(\mathbf{x}) \right] + \nonumber\\ 
&E_{\mathbf{x},\{\mathbf{y}_1,...,\mathbf{y}_K\}}[ \log\left( 
1 - D^{c_y}\big{(}\bar{\mathbf{x}}\right)]
\end{align}	
The superscript attached to $D$ denotes the object class; the loss is computed only using the corresponding binary prediction score of the class. 	
	
The content reconstruction loss helps $G$ learn a translation model. Specifically, when using the same image for both the input content image and the input class image (in this case $K=1$), the loss encourages $G$ to generate an output image identical to the input
\begin{align}
\mathcal{L}_{\text{R}}(G) = E_{\mathbf{x}}\left[||\mathbf{x}-G(\mathbf{x},\{\mathbf{x}\})||_1^1\right].
\end{align}	

The feature matching loss regularizes the training. We first construct a feature extractor, referred to as $D_f$, by removing the last (prediction) layer from $D$. We then use $D_f$ to extract features from the translation output $\bar{\mathbf{x}}$ and the class images $\{\mathbf{y}_1,...,\mathbf{y}_K\}$ and minimize 
\begin{align}
\mathcal{L}_{\text{F}}(G) = E_{\mathbf{x},\{\mathbf{y}_1,...,\mathbf{y}_K\}}\big[||D_f(\bar{\mathbf{x}}))-\sum_k \frac{D_f (\mathbf{y}_k)}{K}||_1^1 \big].
\end{align}

Both of the content reconstruction loss and the feature matching loss are not new topics to  image-to-image translation~\cite{liu2017unsupervised,huang2018multimodal,wang2018high,park2019semantic}. Our contribution is in extending their use to the more challenging and novel few-shot unsupervised image-to-image translation setting.

\begin{figure*}[bth!]	
	\centering
	\includegraphics[trim=0.00in 0.0in 0.0in 0in,width=0.99\textwidth]{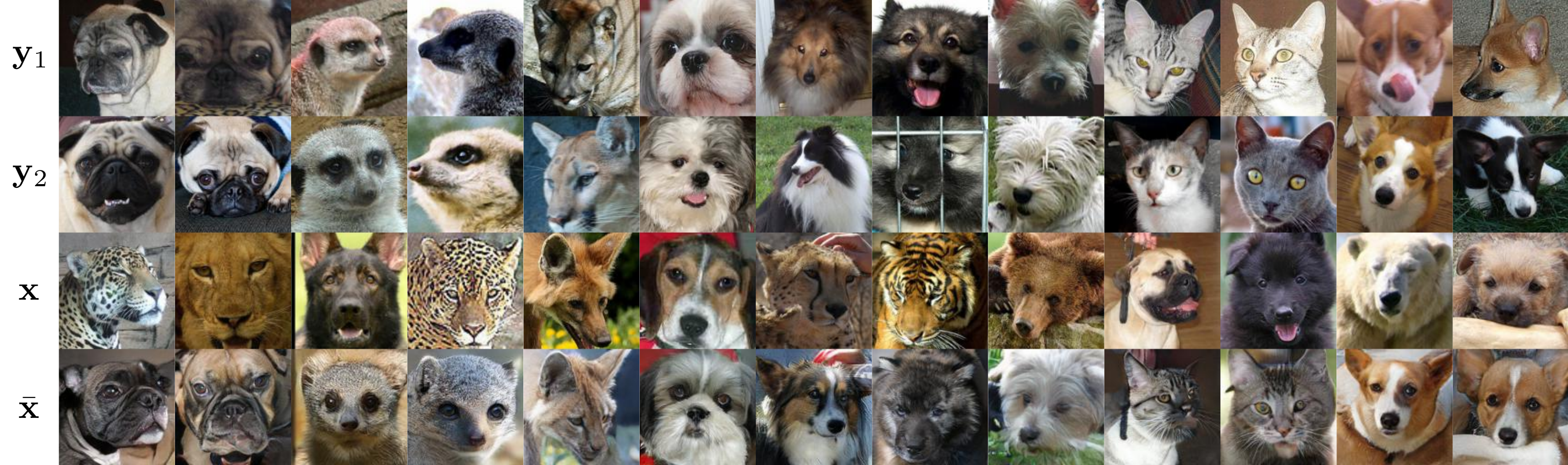}
	\includegraphics[trim=0.00in 0.0in 0.0in 0in,width=0.99\textwidth]{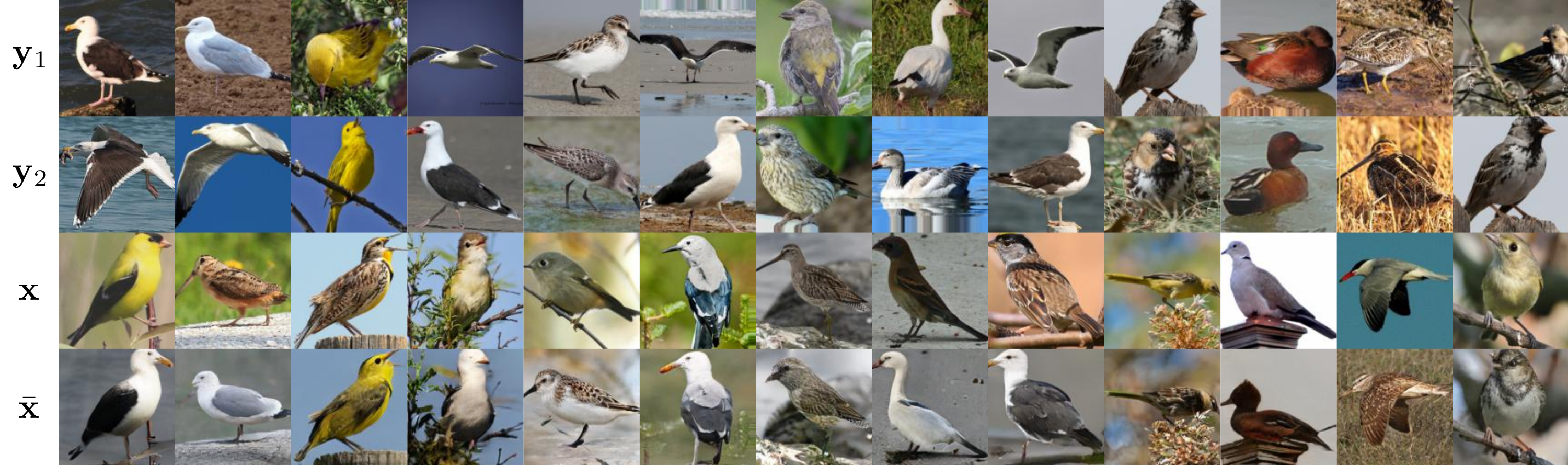}
	\includegraphics[trim=0.00in 0.0in 0.0in 0in,width=0.99\textwidth]{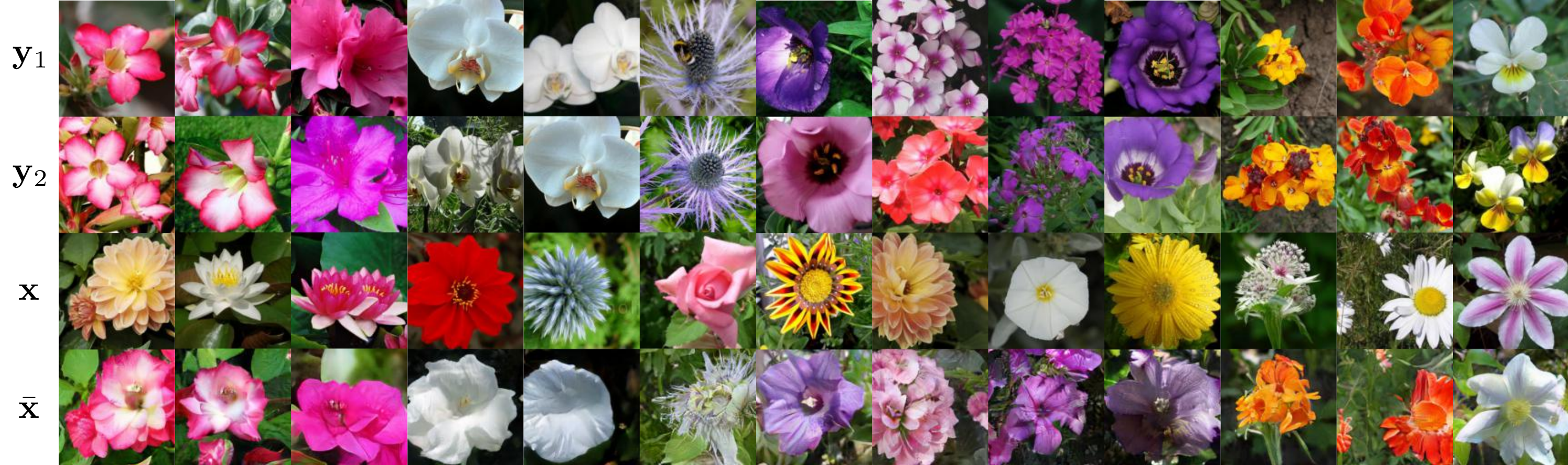}	
	\includegraphics[trim=0.00in 0.0in 0.0in 0in,width=0.99\textwidth]{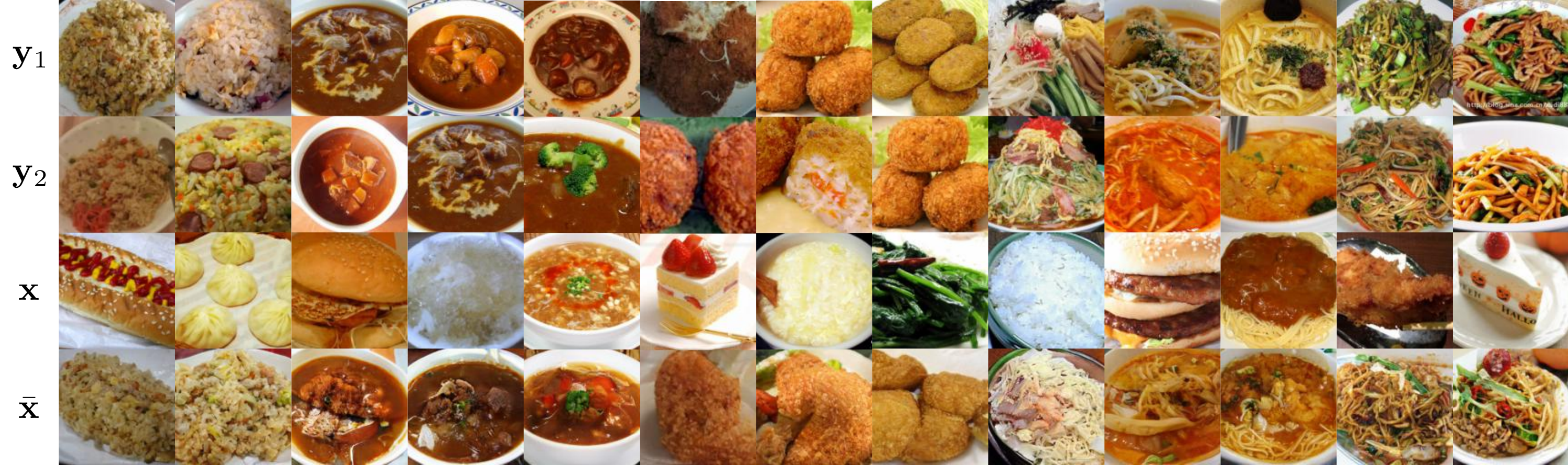}	
	\scaption{Visualization of the \textit{few-shot} unsupervised image-to-image translation results. The results are computed using the \texttt{FUNIT-5} model. From top to bottom, we have the results from the animal face, bird, flower, and food datasets. We train one model for each dataset. For each example, we visualize 2 out of 5 randomly sampled class images $\mathbf{y}_1 \mathbf{y}_2$, the input content image $\mathbf{x}$, and the translation output $\bar{\mathbf{x}}$. The results show that \proposed generate plausible translation outputs under the difficult few-shot setting where the models see no images from any of the target classes during training. We note that the objects in the output images have similar poses to the inputs. }
	\label{fig::vis_results}\vspace{-2mm}
\end{figure*}

\section{Experiments}
\label{sec:expts}

\mysubsection{Implementation.} We set $\lambda_{\text{R}}=0.1$ and $\lambda_{\text{F}}=1$. We optimize (\ref{eqn::learning}) using RMSProp with learning rate $0.0001$. We use the hinge version of GAN loss~\cite{lim2017geometric,miyato2018spectral,zhang2018self,brock2018large} and the real gradient penalty regularization proposed by Mescheder \etal\cite{mescheder2018training}. The final generator is a historical average version of the intermediate generators~\cite{karras2018progressive} where the update weight is $0.001$. We train the \proposed model using $K=1$ since we desire it to work well even when \textit{only one} target class image is available at test time. In the experiments, we evaluate its performance under $K=1,5,10,15,20$. Each training batch consists of 64 content images, which are evenly distributed on 8 V100 GPUs in an NVIDIA DGX1 machine.

\mysubsection{Datasets.} We use the following datasets for experiments.

\begin{itemize}[leftmargin=*,noitemsep,topsep=0pt]	
	\item {\it Animal Faces.} We build this dataset using images from the 149 carnivorous animal classes in ImageNet~\cite{deng2009imagenet}. We first manually label bounding boxes of 10000 carnivorous animal faces in the images. We then train a Faster RCNN~\cite{girshick2015fast} to detect animal faces in the images. We only use the bounding boxes with high detection scores. This renders a set of 117574 animal faces. We split the classes into a source class set and a target class set, which contains 119 and 30 animal classes, respectively. 
	\item {\it Birds}~\cite{van2015building}. 48527 images of 555 North American bird species. 444 species are used for the source class set and 111 species are used for the target class set.
	\item {\it Flowers}~\cite{nilsback2008automated}. 8189 images from 102 species. The source and target sets have 85 and 17 species, respectively.	
	\item {\it Foods}~\cite{Kawano2014automatic}. 31395 images from 256 kinds of food. The source and target set have 224 and 32 kinds, respectively.	
\end{itemize}

\mysubsection{Baselines.} Depending on whether images of the target class are available during training, we define two sets of baselines: \textit{fair} (unavailable) and \textit{unfair} (available). 

\begin{itemize}[leftmargin=*,topsep=2pt]	
	\item {\it Fair.} This is the setting of the proposed \proposed framework. As none of the prior unsupervised image-to-image translation methods are designed for the setting, we build a baseline by extending the StarGAN method~\cite{choi2017stargan}, which is the state of the art for multi-class unsupervised image-to-image translation. We train a StarGAN model purely using source class images. During testing, given $K$ images of a target class, we compute the average VGG~\cite{simonyan2015very} Conv5 features for the $K$ images and compute its cosine distance to the average VGG Conv5 feature for the images of each source class. We then compute the class association vector by applying softmax to the cosine distances. We use the class association vector as input to the StarGAN model (substituting the one-hot class association vector input) for generating images of unseen target classes. The baseline method is designed with the assumption that the class association scores could encode how an unseen target object class is related to each of the source classes, which can be used for few-shot generation. We denote this baseline \texttt{StarGAN-Fair-K}.
	
	\item {\it Unfair.} These baselines include target class images in the training. We vary the number of available images ($K$) per target class from 1 to 20 and train various unsupervised image-to-image translation models. We denote the StarGAN model that is trained with $K$ images per target class as \texttt{StarGAN-Unfair-K}. We also train several state-of-the-art two-domain translation models including CycleGAN~\cite{zhu2017unpaired}, UNIT~\cite{liu2017unsupervised}, and MUNIT~\cite{huang2018multimodal}. For them, we treat images of the source classes as the first domain and images of one target class as the second domain. This results in $|\mathbb{T}|$ unsupervised image-to-image translation models per dataset per two-class baseline. We label these baselines as \texttt{CycleGAN-Unfair-K}, \texttt{UNIT-Unfair-K}, and \texttt{MUNIT-Unfair-K}.
\end{itemize}
For the baseline methods, we use the source code and default parameter settings provided by the authors.

\mysubsection{Evaluation protocol.} We use a randomly sampled 25000 images from the source classes as the content images. We then translate them to each target class by randomly sample $K$ images of the target class. This produces $|\mathbb{T}|$ sets of images for each competing approach and they are used for evaluation. We use the same $K$ images for each content image for all the competing approaches. We test a range of $K$ values, including 1, 5, 10, 15, and 20. 

\begin{figure}[!tb]
	\centering
	\includegraphics[trim=0.00in 0.0in 0in 0in, width=0.49\textwidth]{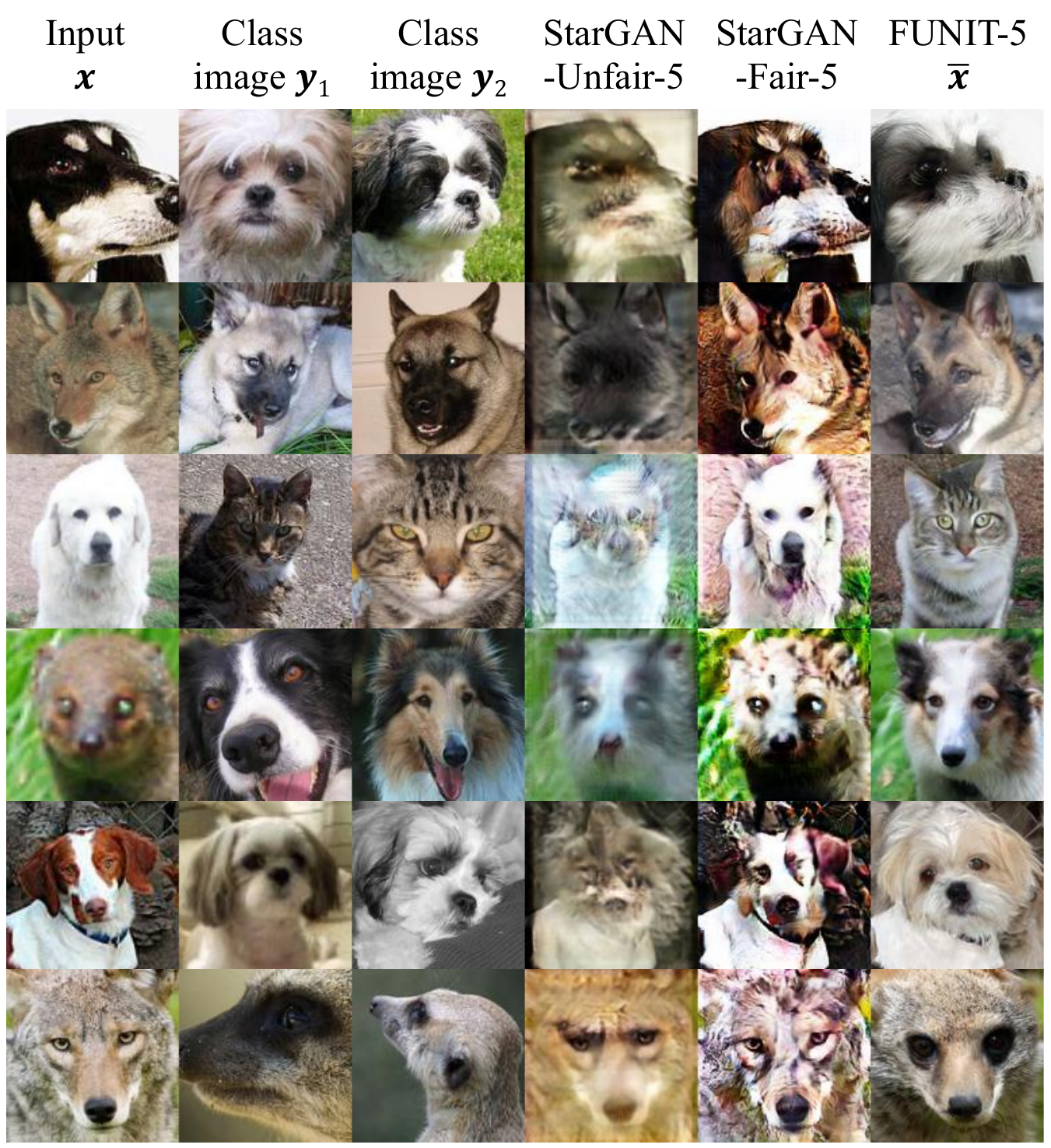}
	\caption{\small Visual comparison of few-shot image-to-image translation performance. From left to right, the columns are input content images $\mathbf{x}$, the two input target class images $\mathbf{y}_1\medspace\mathbf{y}_2$, translation results from the unfair \texttt{StarGAN} baseline, translation results from the fair \texttt{StarGAN} baseline, and results from our framework.}
	\label{fig::vis_comparison}
\end{figure}

\mysubsection{Performance metrics.} We use several criteria for performance comparison. First, we measure whether translations resemble images of the target class. Second, we examine whether class-invariant content are preserved during translation. Third, we quantify photorealism of output images. Finally, we measure whether the model can be used to generate the image distribution of a target class. More details on the performance metrics are given in Appendix~\ref{sec::metric}.
\begin{itemize}[leftmargin=*,topsep=2pt]
	\item {\it Translation accuracy} measure whether a translation output belongs to the target class. We use two Inception-V3~\cite{szegedy2016rethinking} classifiers. One classifier is trained using both of the source and target classes (denoted as \textit{all}), while the other is trained using the target classes along (denoted as \textit{test}). We report both Top1 and Top5 accuracies.
	
	\item {\it Content preservation} is based on a variant of perceptual distance~\cite{johnson2016perceptual,zhang2018unreasonable}, called the domain-invariant perceptual distance (DIPD)~\cite{huang2018multimodal}. The distance is given by L2 distance between two normalized VGG~\cite{simonyan2015very} Conv5 features, which is more invariant against domain change~\cite{huang2018multimodal}.
	
	\item {\it Photorealism.} This is measured by the inception scores (IS)~\cite{salimans2016improved}. We report inception scores using the two inception classifiers trained for measuring translation accuracy, denoted by \textit{all} and \textit{test}, respectively. 
	
	\item {\it Distribution matching} is based on Fr\'echet Inception Distance (FID)~\cite{heusel2017gans}. We compute FID for each of the $|\mathbb{T}|$ target object classes and report their mean FID (mFID).
\end{itemize}

\mysubsection{Main results.} As shown in Table~\ref{tbl::main_results}, the proposed \proposed framework outperforms the baselines for the few-shot unsupervised image-to-image translation task on all the performance metrics for both the Animal Faces and North American Birds datasets. \proposed achieves 82.36 and 96.05 Top-5 (\textit{test}) accuracy for the 1-shot and 5-shot settings, respectively, on the Animal Face dataset, and 60.19 and 75.75 on the North American Birds dataset. They are all significantly better than those achieved by the corresponding fair baselines. Similar trends can be found for the domain invariant perceptual distance, inception score, and Fr\'echet inception distance. Moreover, with just 5 shots, \texttt{FUNIT} outperforms all the unfair baselines under 20-shot settings. Note that for the results of \texttt{CycleGAN-Unfair-20}, \texttt{UNIT-Unfair-20}, and \texttt{MUNIT-Unfair-20} are from $|\mathbb{T}|$ image-to-image translation networks, while our method is from a single translation network.

The table also shows that the performance of the proposed \proposed model is positively correlated with number of available target images $K$ at test time. A larger $K$ leads to improvements across all the metrics, and the largest performance boost comes from $K=1$ to $K=5$. The \texttt{StarGAN-Fair} baseline does not exhibit a similar trend.

In Figure~\ref{fig::vis_results}, we visualize the few-shot translation results computed by \texttt{FUNIT-5}. The results show that the \proposed model can successfully translate images of source classes to analogous images of novel classes. The poses of the object in the input content image $\mathbf{x}$ and the corresponding output image $\bar{\mathbf{x}}$ remain largely the same. 
The output images are photorealistic and resemble images from the target classes. More results are given in Appendix~\ref{sec::more}.

In Figure~\ref{fig::vis_comparison}, we provide a visual comparison. As the baselines are not designed for the few-shot image translation setting, they failed in the challenging translation task. They either generate images with a large amount of artifacts or just output the input content image. On the other hand, \proposed generates high-quality image translation outputs. 

\mysubsection{User study.} To compare the photorealism and faithfulness of the translation outputs, we perform human evaluation using the Amazon Mechanical Turk~(AMT) platform. Specifically, we give the workers a target class image and two translation outputs from different methods~\cite{wang2018high,huang2018multimodal} and ask them to choose the output image that resembles more the target class image. The workers are given unlimited time to make the selection. We use both the Animal Faces and North American Birds datasets. For each comparison, we randomly generate $500$ questions and each question is answered by $5$ different workers. For quality control, a worker must have a lifetime task approval rate grater than 98\% to be able to participate in the evaluation.

\begin{figure*}[!bht]
	\centering
	\includegraphics[trim=0.00in 0.0in 0.3in 0.6in,width=0.245\textwidth]{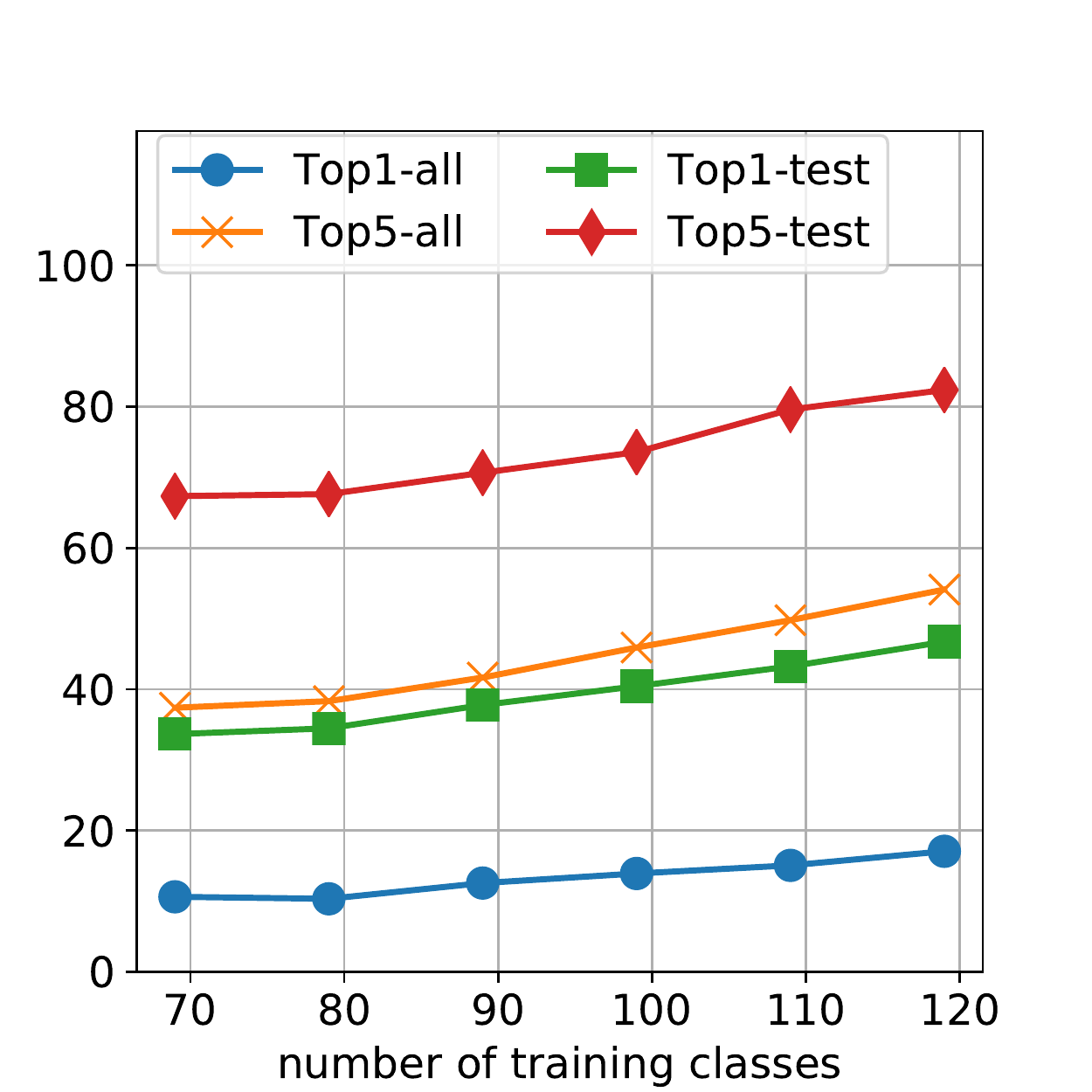}
	\includegraphics[trim=0.00in 0.0in 0.3in 0.6in,width=0.245\textwidth]{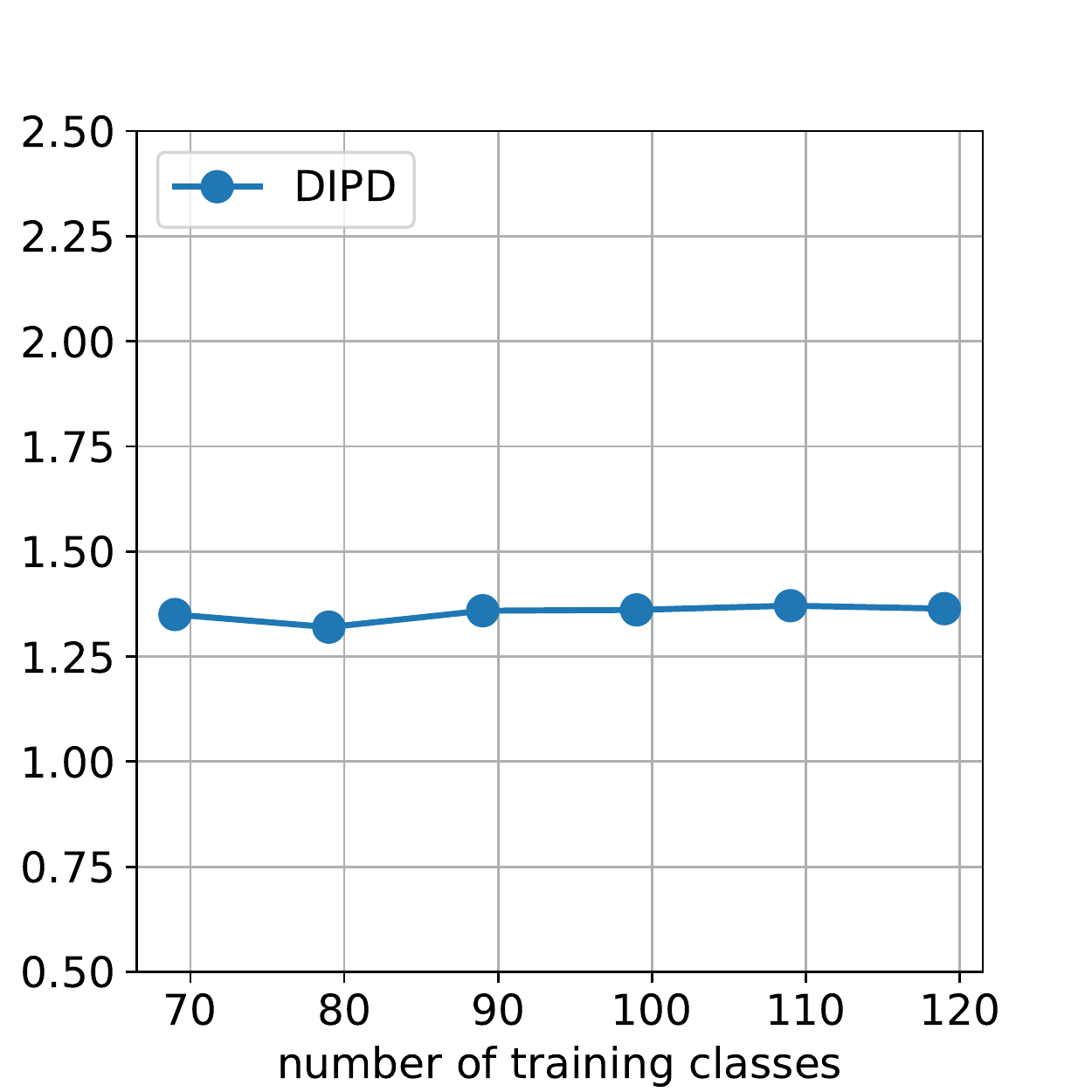}
	\includegraphics[trim=0.00in 0.0in 0.3in 0.6in,width=0.245\textwidth]{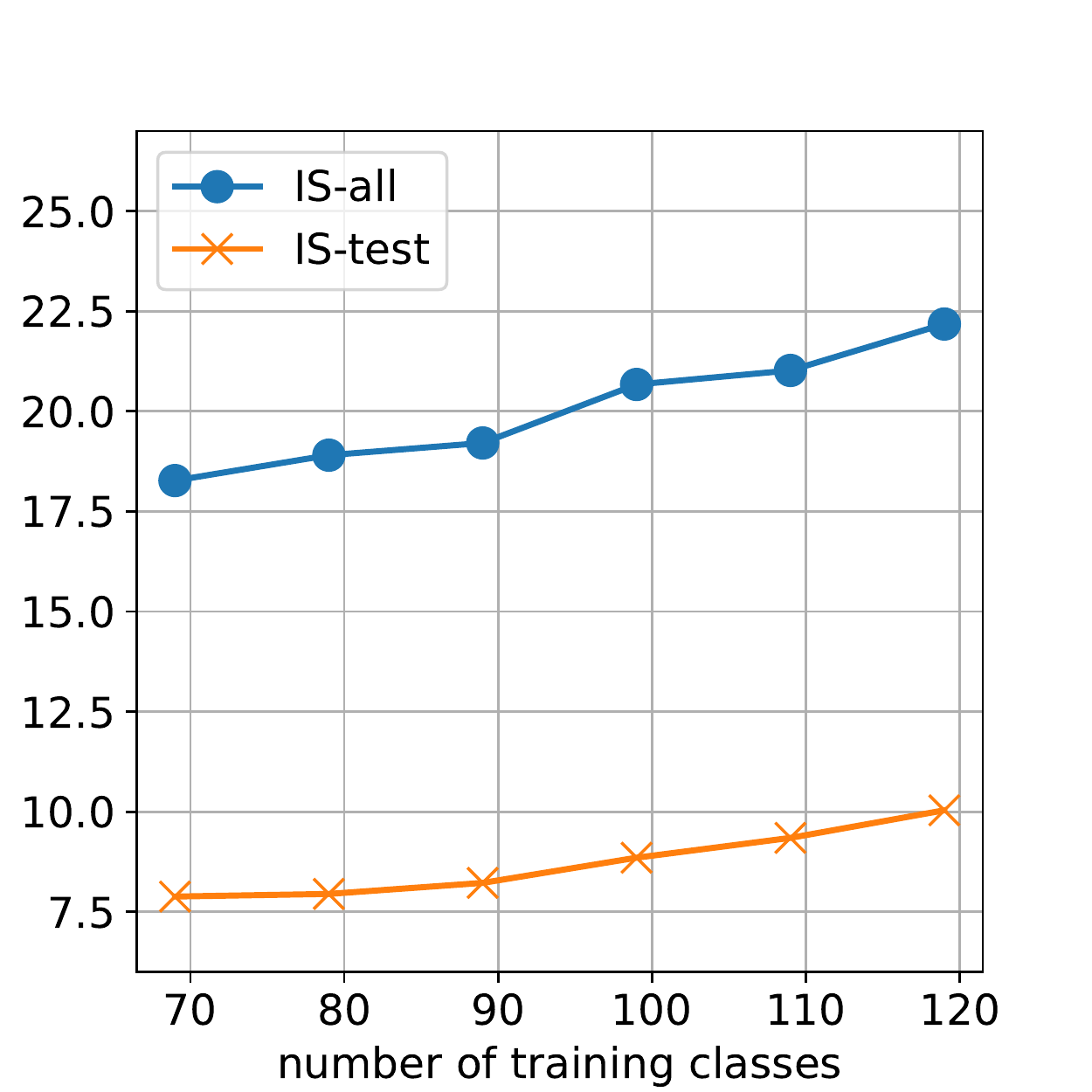}
	\includegraphics[trim=0.00in 0.0in 0.3in 0.6in,width=0.245\textwidth]{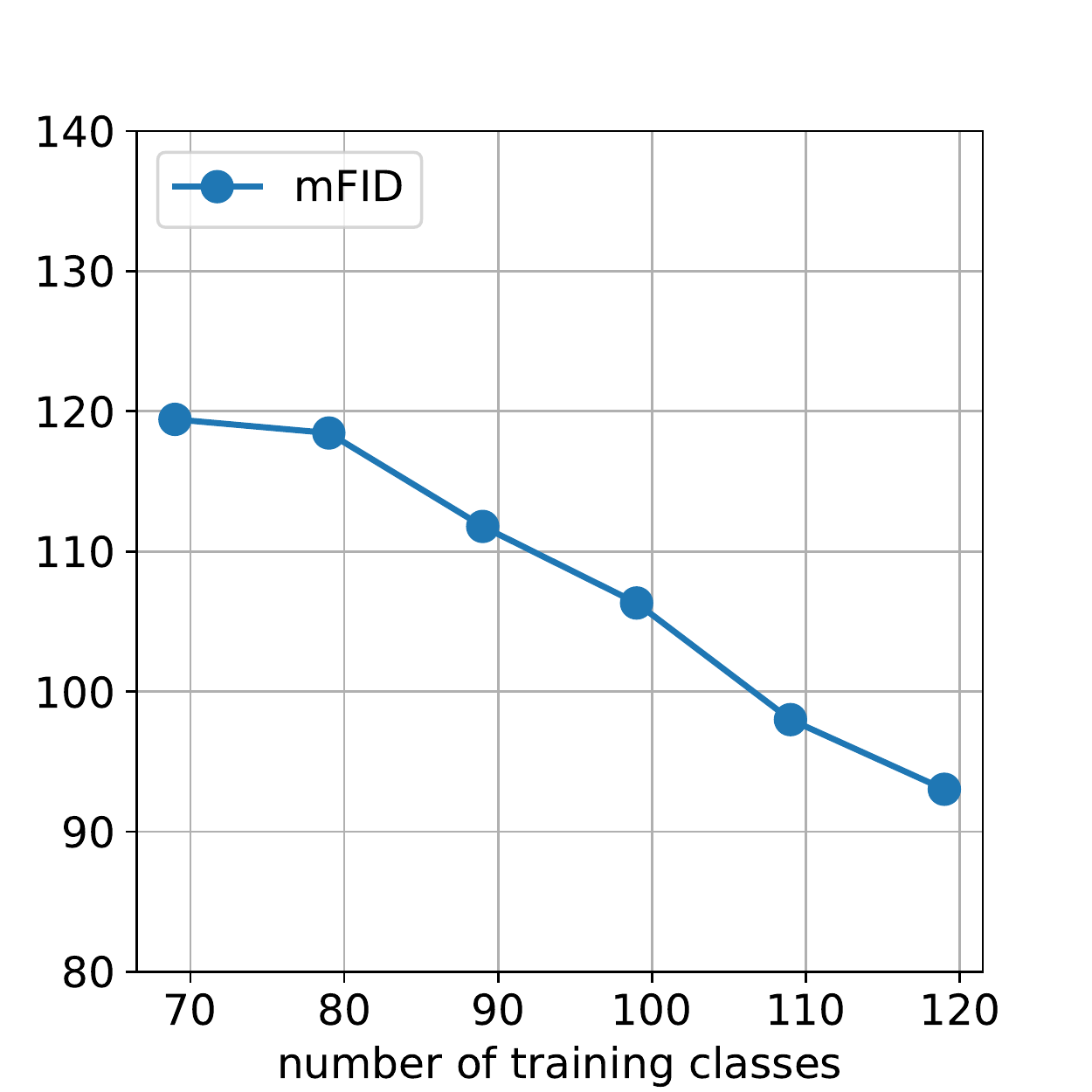}
	\vspace{1mm}
	\scaption{Few-shot image translation performance vs. number of object classes seen during training on the Animal Faces dataset. The performance is positively correlated with number of source object classes seen during training.}
	\label{fig::object_class}
\end{figure*}
\begin{table}[t!]
	\centering
	\small
\begin{tabular}{|l|c|c|}
		\hline
		\multicolumn{1}{|c|}{\bf Setting} & \bf Animal & \bf Birds \\
		\hline		
		\texttt{FUNIT-5} vs. \texttt{StarGAN-Fair-5} & 86.08 & 82.56 \\
		\texttt{FUNIT-5} vs. \texttt{StarGAN-Unfair-20} & 86.00 & 84.48 \\
		\texttt{FUNIT-5} vs. \texttt{CycleGAN-Unfair-20} & 71.68 & 77.76 \\
		\texttt{FUNIT-5} vs. \texttt{UNIT-Unfair-20} & 77.84 & 77.96 \\
		\texttt{FUNIT-5} vs. \texttt{MUNIT-Unfair-20} & 83.56 & 79.64 \\		
		\hline		
	\end{tabular}
	\vspace{0.5mm}
	\scaption{User preference score. The numbers indicate the percentage of users favors results generated by the proposed method over those generated by the competing method.}\label{tbl::user_study}		
 \end{table}
 
 \begin{table}[t!]
	\centering
	\small
	\resizebox{\columnwidth}{!}{%
		\begin{tabular}{|c|C{1.3cm}|C{1.3cm}||C{1.3cm}|C{1.3cm}|}
			\hline
			\bf \# of generated & \multicolumn{2}{c||}{\bf Animal Face} & \multicolumn{2}{c|}{\bf North American Birds} \\
			\bf samples $N$ & \bf \texttt{S\&H}~\cite{hariharan2017low} & \bf \proposed & \bf \texttt{S\&H}~\cite{hariharan2017low} & \bf \proposed \\
			\hline
			0 & \multicolumn{2}{c||}{38.76} & \multicolumn{2}{c|}{30.38} \\
			10 & 40.51 & {\bf 42.05} & 31.77 & {\bf 33.41} \\
			50 & 40.24 & {\bf 42.22} & 31.66 & {\bf 33.64} \\
			100 & 40.76 & {\bf 42.14} & 32.12 & {\bf 34.39} \\
			\hline
		\end{tabular}
	}
	\vspace{0.5mm}
	\scaption{Few-shot classification accuracies averaged over 5 splits.}	\label{tbl::few_shot_classification}	
\end{table}
\begin{figure}[bth!]	
	\centering
	\includegraphics[trim=0.00in 0.0in 0.0in 0in,width=0.49\textwidth]{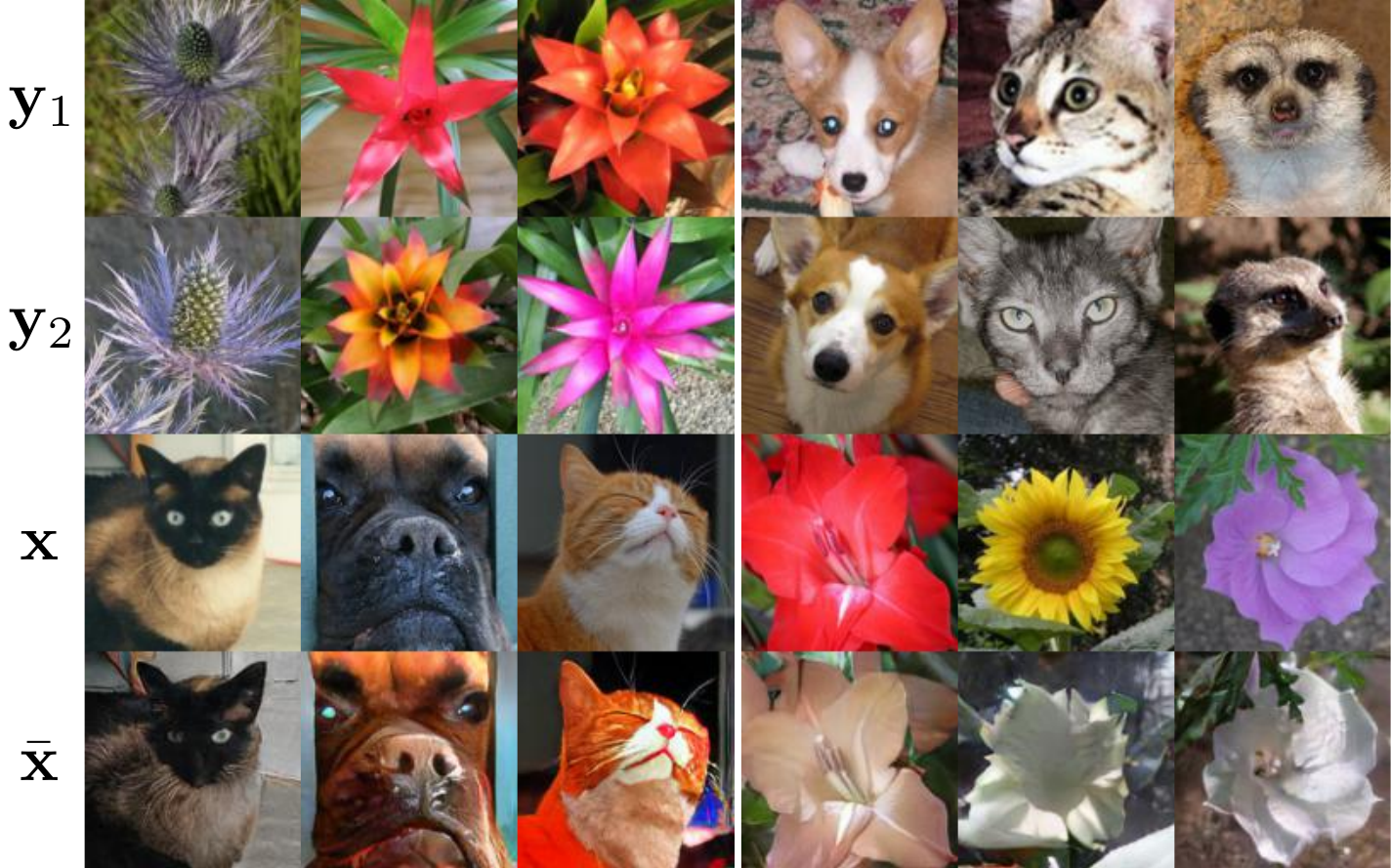}
	\scaption{Limitations of the proposed framework. When the appearance of a unseen object class is dramatically different to the appearances of the source classes, (\eg flower and animal face). The proposed \proposed framework fails to generate meaningful translation outputs.}
	\label{fig::vis_mixing}
\end{figure}

According to Table~\ref{tbl::user_study}, the human subjects consider the translation outputs generated by the proposed method under the 5-shot setting (\texttt{FUNIT-5}) much more similar to the target class images than those generate by the fair baseline under the same setting (\texttt{StarGAN-Fair-5}). Even when compared with the results of unfair baselines that have access to 20 images per target class at training time, our translation results are still considered to be much more faithful.

\mysubsection{Number of source classes in the training set.} In Figure~\ref{fig::object_class}, we analyze the performance versus varying number of source classes in the training set under the one-shot setting (\texttt{FUNIT-1}), using the animal dataset. We plot the curves by varying the number from 69 to 119 classes with an interval of 10. As shown, the performance is positively correlated with the number of object classes in terms of translation accuracy, image quality, and distribution matching. The domain-invariant perceptual distance remains flat. This shows that a \proposed model that sees more object classes (larger diversity) during training performs better during testing. A similar trend is observed for the bird dataset, which is given in Appendix~\ref{sec::source}.

\mysubsection{Parameter analysis and ablation study.} We analyze the impact of the individual terms in our objective function and find all of them are essential. Particularly, the content reconstruction loss trades translation accuracy for content preservation score. The details are given in Appendix~\ref{sec::ablation}.

\mysubsection{Comparison with the AdaIN style transfer method.} We train an AdaIN style transfer network~\cite{huang2017adain} for the few-shot animal face translation task and compare the performance as shown in Appendix~\ref{sec::adain}. We find that while the style transfer network can change the textures of the input animals, it does not change their shapes. As a result, the translation outputs still resemble to the inputs. 

\mysubsection{Failure cases.} Several failure cases of the proposed algorithm are visualized in Appendix~\ref{sec::failure}. They include generating hybrid objects, ignoring input content images, and ignoring input class images.

\mysubsection{Latent interpolation.} In Appendix~\ref{sec::latent}, we show interpolation results by keeping the content code fixed while interpolating the class code between two source class images. Interestingly, we find that by interpolating between two source classes (Siamese cat and Tiger) we can sometimes generate a target class (Tabby cat) that the model has never observed. 

\mysubsection{Few-shot translation for few-shot classification.} We evaluate \proposed for few-shot classification using the animal and bird datasets. Specifically, we use the trained \proposed models to generate $N$ (varying from 1, 50, to 100) images for each of the few-shot classes and use the generated images to train the classifiers. We find the classifiers trained with the \proposed generated images consistently achieve better performance than the few-shot classification approach proposed of \texttt{S\&H} by Hariharan~\etal~\cite{hariharan2017low}, which is based on feature hallucination and also has a controllable variable on sample number $N$. The results are shown in Table~\ref{tbl::few_shot_classification} and the experiment details are given in Appendix~\ref{sec::classification}.

\section{Discussion and Future Work}

We introduced the first few-shot unsupervised image-to-image translation framework. We showed that the few-shot generation performance is positively correlated with the number of object classes seen during training and also positively correlated with the number of target class shots provided during test time. 

We provided empirical evidence that \proposed can learn to translate an image of a source class to a corresponding image of an unseen object class by utilizing few example images of the unseen class made available at test time. Although achieving this new capability, \proposed depends on several conditions to work: 1) whether the content encoder $E_x$ can learn a class-invariant latent code $\mathbf{z}_x$, 2) whether the class encoder $E_y$ can learn a class-specific latent code $\mathbf{z}_y$, and, most importantly, 3) whether the class encoder $E_y$ can generalize to images of unseen object classes.

We observed these conditions are easy to meet when the novel classes are visually related to the source classes. However, when the appearance of novel object classes are dramatically different from those of the source classes, \proposed fails to achieve translation as shown in Figure~\ref{fig::vis_mixing}. In this case, \proposed tends to generate color-changed versions of the input content images. This is undesirable but understandable as the appearance distribution has changed dramatically. Addressing this limitation is our future work.

{\small
	\bibliographystyle{ieee_fullname}
	\bibliography{funit}
}
\appendix
\section{Network Architecture}\label{sec::arch}

\begin{figure*}[th!]
	\centering
	\includegraphics[trim=0.00in 0.0in 0in 0in, width=0.99\textwidth]{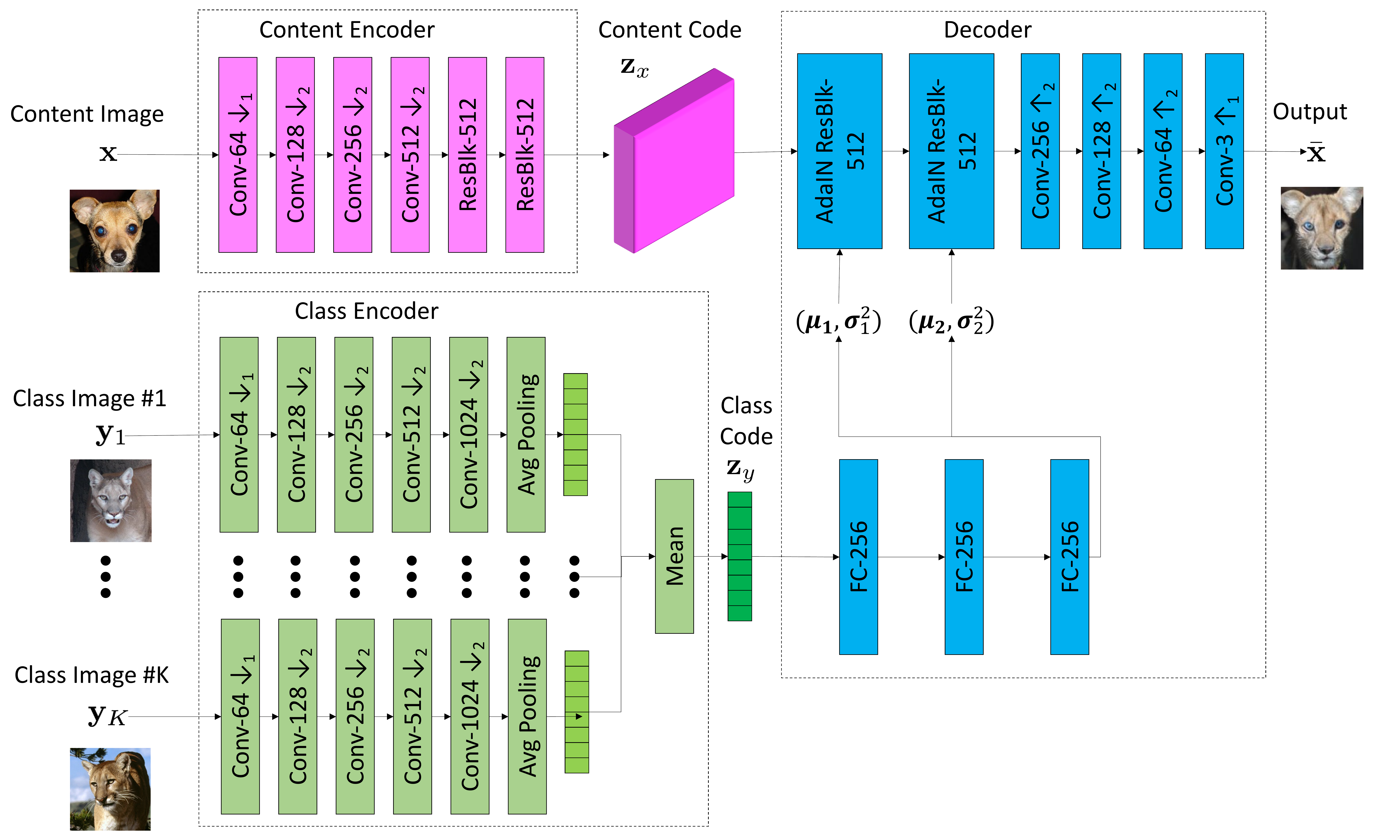}
	\scaption{Visualization of the generator architecture. To generate a translation output $\bar{\pmb{x}}$, the translator combines the class latent code $\pmb{z}_y$ extracted from the class images $\pmb{y}_1,...\pmb{y}_K$ with the content latent code $\pmb{z}_x$ extracted from the input content image. Note that nonlinearity and normalization operations are not included in the visualization.}\label{fig::generator}
\end{figure*}

The few-shot image translator consists of three sub-networks: a content encoder, a class encoder, and a decoder as visualized in Figure~\ref{fig::generator}. The content encoder maps the input content image to a content latent code, which is a feature map. If the resolution of the input image is 128x128, the resolution of the feature map will be 16x16 since there are 3 stride-2 down-sampling operations. This feature map is designed to encode class-invariant content information. It should encode locations of the parts but not their class-specific appearances\footnote{For example, in the animal face translation task, it should encode locations of the ears but not their shape and color.}. On the other hand, the class encoder maps a set of $K$ class images to a class latent code, which is a vector and is aimed to be class-specific. It first maps each input class image to an intermediate latent code using a VGG-like network. These latent vectors are then element-wise averaged to produce the final class latent code.

As shown in the figure, the decoder first decodes the class-specific latent code to a set of mean and variance vectors $(\pmb{\mu}_i,\pmb{\sigma}^2_i)$ where $i=1,2$. These vectors are then used as the affine transformation parameters in the AdaIN residual blocks where $\pmb{\sigma}^2_i$'s are the scaling factors and $\pmb{\mu}_i$'s are the biases. For each residual block, the same affine transformation is applied to every spatial location in the feature map. It controls how the content latent code are decoded to generate the output image. 

The number shown in each block in Figure~\ref{fig::generator} denotes the number of filters in the layer. The nonlinearity and normalization operations included in the network are excluded in the visualization. For the content encoder, each layer is followed by the instance normalization and the ReLU nonlinearity. For the class encoder, each layer is followed by the ReLU nonlinearity. For the decoder, except for the AdaIN residual blocks, each layer is followed by the instance normalization and the ReLU nonlinearity. We upscale the feature maps along each spatial dimension by a factor of 2 using nearest neighbor upsampling.

Our discriminator is a Patch GAN discriminator~\cite{isola2017image}. It utilizes the Leaky ReLU nonlinearity and employs no normalization. The discriminator consists of one convolutional layer followed by 10 activation first residual blocks~\cite{mescheder2018training}. The architecture is illustrated via the following chain of operations: \\
\texttt{Conv-64 $\rightarrow$ ResBlk-128 $\rightarrow$ ResBlk-128 $\rightarrow$ AvePool2x2 $\rightarrow$ ResBlk-256 $\rightarrow$ ResBlk-256 $\rightarrow$ AvePool2x2 $\rightarrow$ ResBlk-512 $\rightarrow$ ResBlk-512 $\rightarrow$ AvePool2x2 $\rightarrow$ ResBlk-1024 $\rightarrow$ ResBlk-1024 $\rightarrow$ AvePool2x2 $\rightarrow$ ResBlk-1024 $\rightarrow$ ResBlk-1024 $\rightarrow$ Conv-$||\mathbb{S}||$} where $||\mathbb{S}||$ is the number of source classes.

\section{Performance Metrics}\label{sec::metric}

\mysubsection{Translation accuracy.} We use two Inception-V3~\cite{szegedy2016rethinking} classifiers to measure translation accuracy. The first classifier, denoted as \textit{all}, is obtained by finetuning the ImageNet-pretrained Inception-V3 model on the task of classifying all the source and target object classes (\eg all of the 149 classes for the Animal Faces dataset and all of the 555 classes for the North American Birds dataset). The second classifier (denoted as \textit{test}) is obtained by finetuning the ImageNet-pretrained Inception-V3 model on the task of classifying the target object classes (\eg 30 target classes for the Animal Faces dataset and 111 target classes for the North American Birds dataset). We apply the classifiers to the translation output to see if they can recognize the output as an image of the target class. If yes, we denote it as a correct translation. We compare performance of competing models using both Top1 and Top5 accuracies. We thus have 4 evaluation metrics for translation accuracy: \textit{Top1-all}, \textit{Top5-all}, \textit{Top1-test}, and \textit{Top5-test}. An unsupervised image-to-image translation model with a higher accuracy is better. We note that similar evaluation protocols were used for comparing image-to-image translation models on the semantic label map to image translation  task~\cite{isola2017image,wang2018high,chen2017photographic}.
	
\mysubsection{Content preservation.} We quantify the content preservation performance using the domain-invariant perceptual distance (DIPD)~\cite{huang2018multimodal}. The DIPD is a variant of perceptual distance~\cite{johnson2016perceptual,zhang2018unreasonable}. To compute the DIPD, we first extract the VGG~\cite{simonyan2015very} conv5 feature from the input content image as well as from the output translation image. We then apply the instance normalization~\cite{ulyanov2017improved} to the features, which will remove their mean and variance. This way, we can filter out much class-specific information in the features~\cite{huang2017adain,li2016revisiting} and focus on the class-invariant similarity. The DIPD is given by L2 distance between the instance normalized features.
	
\mysubsection{Photorealism.} We use the inception score (IS)~\cite{salimans2016improved}, which is widely used for quantifying image generation performance. Let $p(t|\mathbf{y})$ be the distribution of class label $t$ of the inception model over the output translation image $\mathbf{y}$. The inception score is given by 
\begin{align}
\text{IS}_C = \exp(
\text{E}_{\mathbf{y}\sim p(\mathbf{y})}[\text{KL}(p(t|\mathbf{y})|p(t))])
\label{equ:is}
\end{align} 
where $p(t)=\int_{\mathbf{y}} p(t|\mathbf{y}) d\mathbf{y}$. It is argued in Salimans \etal~\cite{salimans2016improved} that the inception score is positively correlated with visual quality of neural network-generated images.
	
\mysubsection{Distribution matching.} The Frechet Inception Distance FID~\cite{heusel2017gans} is designed for measuring similarities between two sets of images. We use the activations from the last average pooling layer of the ImageNet-pretrained Inception-V3~\cite{szegedy2016rethinking} model as the feature vector of an image for computing FID. As we have $|\mathbb{T}|$ unseen classes, we translate source images to each of the $|\mathbb{T}|$ unseen classes and produce $|\mathbb{T}|$ sets of translation outputs. For each of the $|\mathbb{T}|$ sets of translation outputs, we compute the FID between the set to the corresponding set of ground truth images. This renders $|\mathbb{T}|$ FID scores. The average of the $|\mathbb{T}|$ FID scores is used as our final distribution matching performance metric, which is referred to as the mean FID (mFID). 

\section{Effect on Number of Source Classes}\label{sec::source}

\begin{figure*}[bth!]
	\centering
	\includegraphics[trim=0.00in 0.0in 0.3in 0in,width=0.245\textwidth]{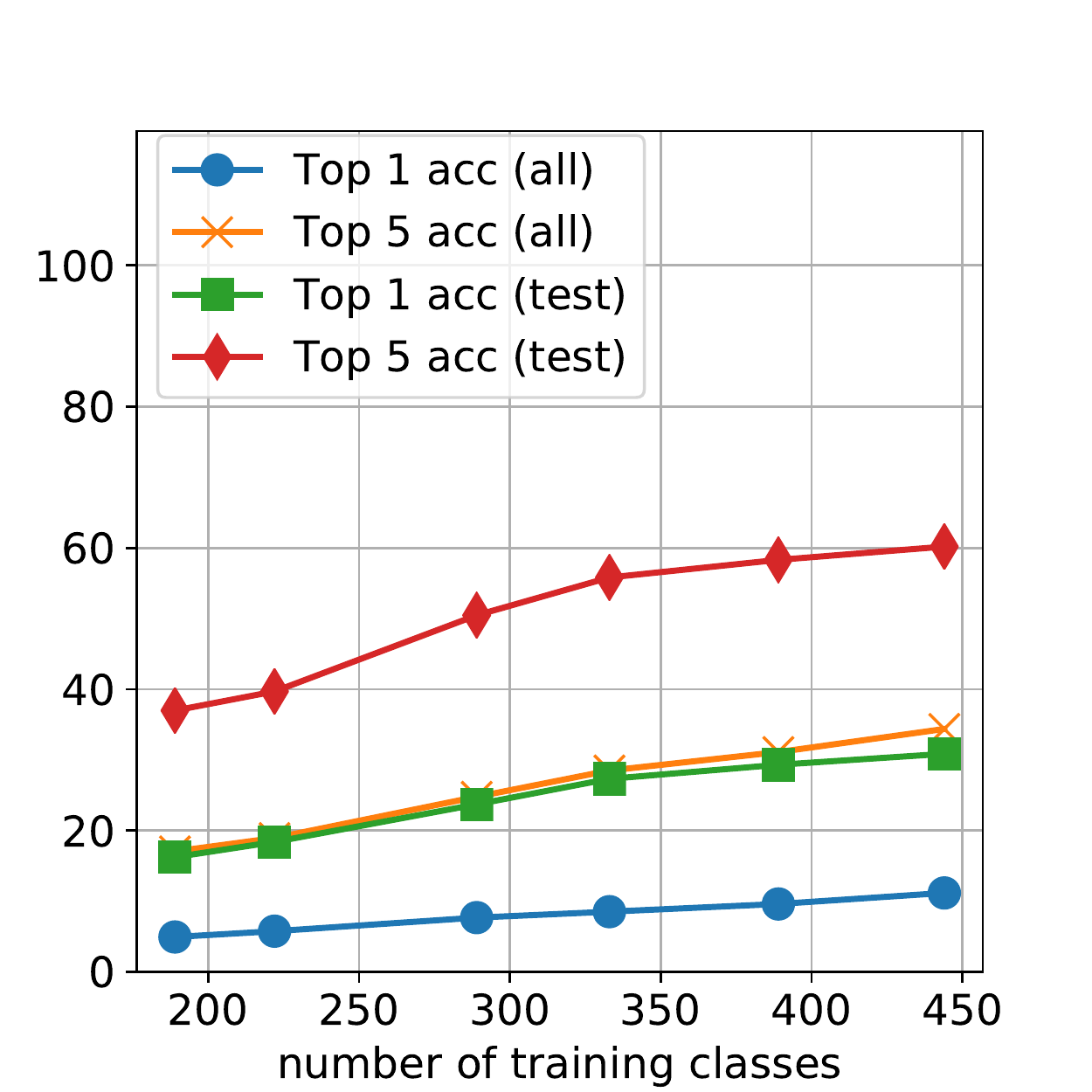}
	\includegraphics[trim=0.00in 0.0in 0.3in 0in,width=0.245\textwidth]{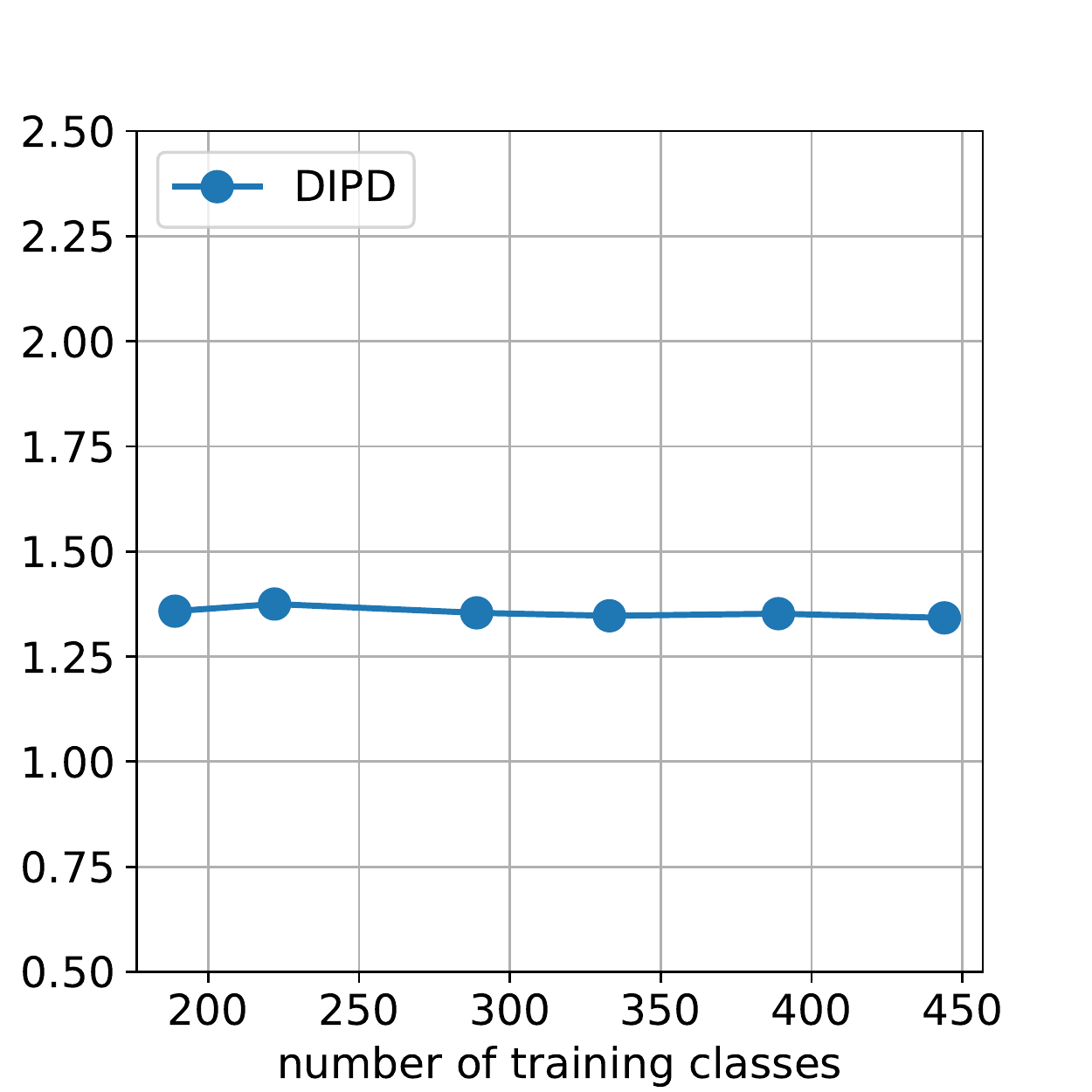}
	\includegraphics[trim=0.00in 0.0in 0.3in 0in,width=0.245\textwidth]{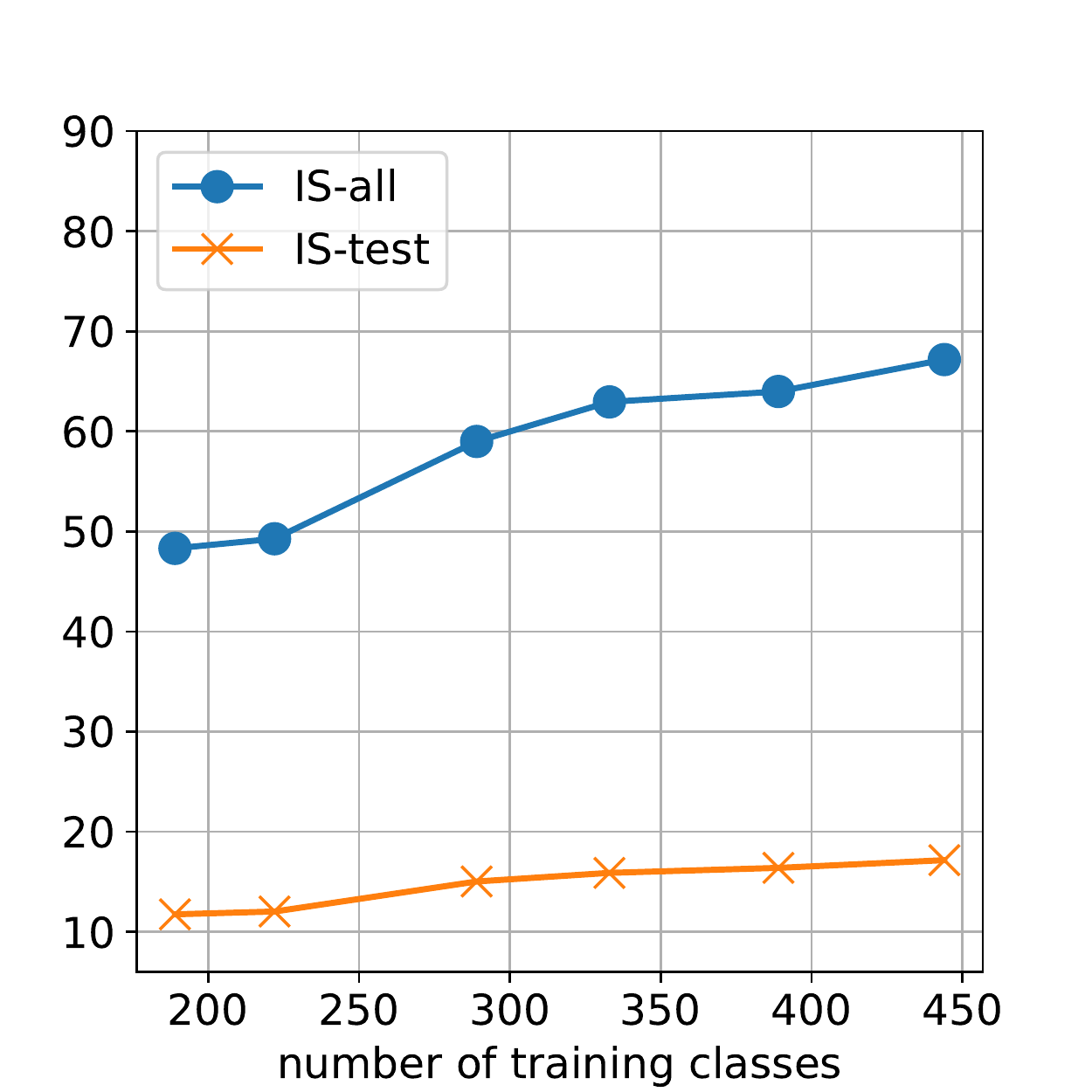}
	\includegraphics[trim=0.00in 0.0in 0.3in 0in,width=0.245\textwidth]{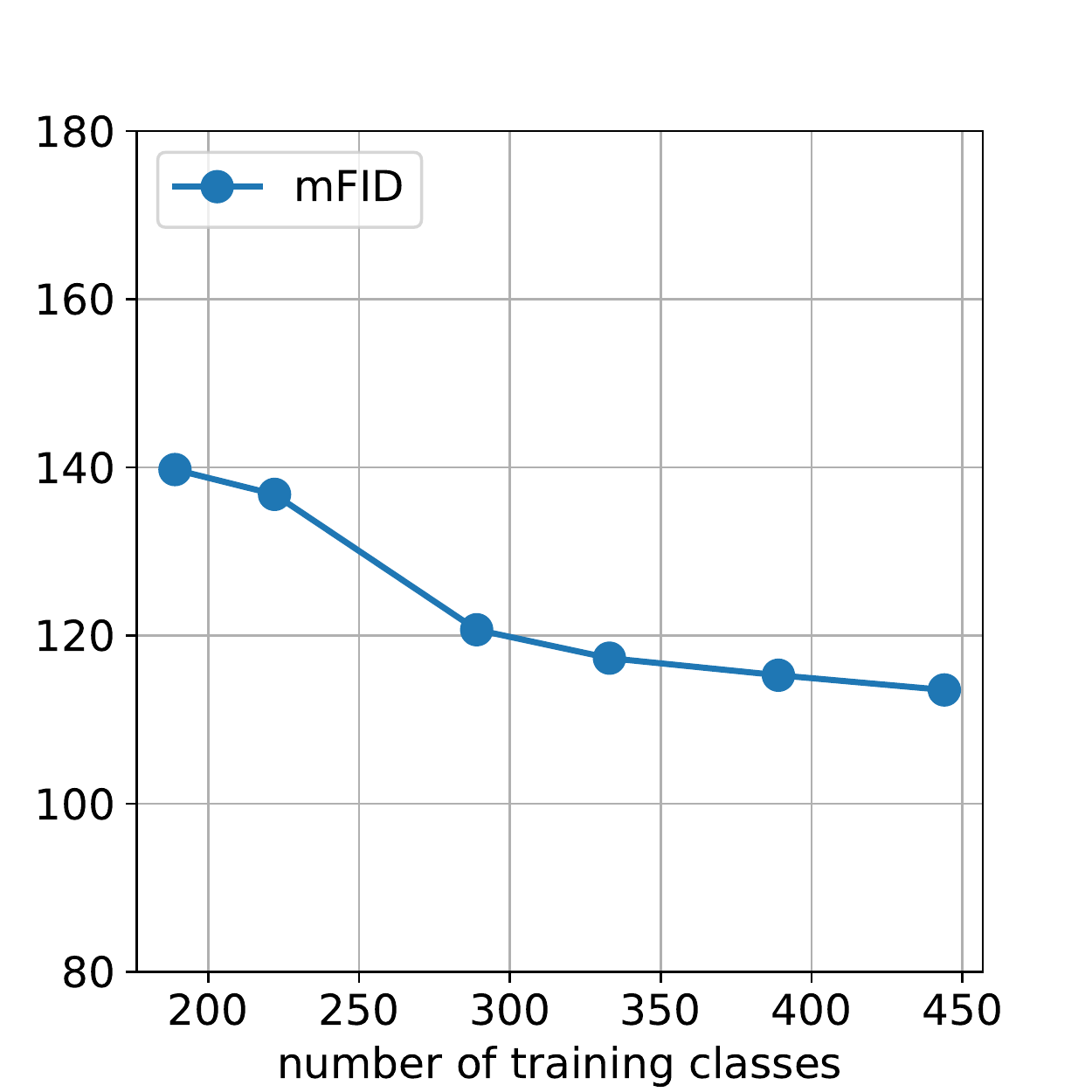}	
	\scaption{Few-shot image translation performance vs. number of object classes seen during training on the North American Birds dataset.}
	\label{fig::object_class_nabirds}
\end{figure*}

\begin{table*}[bth!]
	\centering	
	\small		
	{\tabcolsep=6pt\def\arraystretch{1}
		\begin{tabular}{|c||cccc|c|cc|c|}
			\hline
			Setting & \bf Top1-all $\uparrow$ & \bf Top5-all $\uparrow$ & \bf Top1-test $\uparrow$ & \bf Top5-test $\uparrow$ & \bf DIPD $\downarrow$& \bf IS-all $\uparrow$ & \bf IS-test $\uparrow$ & \bf mFID $\downarrow$\\
			\hline
			$\lambda_{\text{R}}=0.01$& 16.02&52.30&45.52&81.68&1.370&21.80&9.73&94.98\\
			$\lambda_{\text{R}}=0.1$& \textbf{17.07}&\textbf{54.11}&\textbf{46.72}&\textbf{82.36}&1.364&22.18&\textbf{10.04}&\textbf{93.03}\\
			$\lambda_{\text{R}}=1$& 16.60&52.05&45.62&81.77&1.346&\textbf{22.21}&9.81&94.23\\
			$\lambda_{\text{R}}=10$& 13.04&44.32&39.06&75.81&\textbf{1.298}&20.48&8.90&108.71\\
			\hline
		\end{tabular}}\vspace{1mm}
	\scaption{Parameter sensitivity analysis on the content image reconstruction loss weight,  $\lambda_{\text{R}}$. $\uparrow$ means larger numbers are better, $\downarrow$ means smaller numbers are better. The value of 0.1 provides a good trade-off between content preservation and translation accuracy, which is used as the default value throughout the paper. We use the \texttt{FUNIT-1} model for this experiment. }\label{tbl::lambda_r}
\end{table*}
\begin{table*}[bth!]
\centering	
\small		
{\tabcolsep=6pt\def\arraystretch{1}
	\begin{tabular}{|c||cccc|c|cc|c|}
	\hline
	Setting & \bf Top1-all $\uparrow$ & \bf Top5-all $\uparrow$ & \bf Top1-test $\uparrow$ & \bf Top5-test $\uparrow$ & \bf DIPD $\downarrow$& \bf IS-all $\uparrow$ & \bf IS-test $\uparrow$ & \bf mFID $\downarrow$\\
	\hline
		\st{FM} &15.33&52.98&46.33&\textbf{82.43}&1.401&\textbf{22.45}&9.86&\textbf{92.98}\\
		\st{GP} &1.15&4.74&3.18&15.50&1.752&1.78&1.84&316.56\\
		\ \ proposed\ \  &\textbf{17.07}&\textbf{54.1}1&\textbf{46.72}&82.36&\bf{1.364}&22.18&\textbf{10.04}&93.03\\
		\hline
	\end{tabular}}\vspace{1mm}		
\scaption{Ablation study on the object terms. $\uparrow$ means larger numbers are better, $\downarrow$ means smaller numbers are better. \st{FM} represents a setting of the proposed framework with the feature matching loss term removed, while \st{GP} represents a setting of the proposed framework without the gradient penalty loss. The default setting renders better performances on various criteria most of the time. We use the \texttt{FUNIT-1} model for this experiment.}\label{tbl::ablation}	
\end{table*}
\begin{figure*}[bth!]
	\centering
	\includegraphics[trim=0.00in 0.0in 0.3in 0in,width=0.24\textwidth]{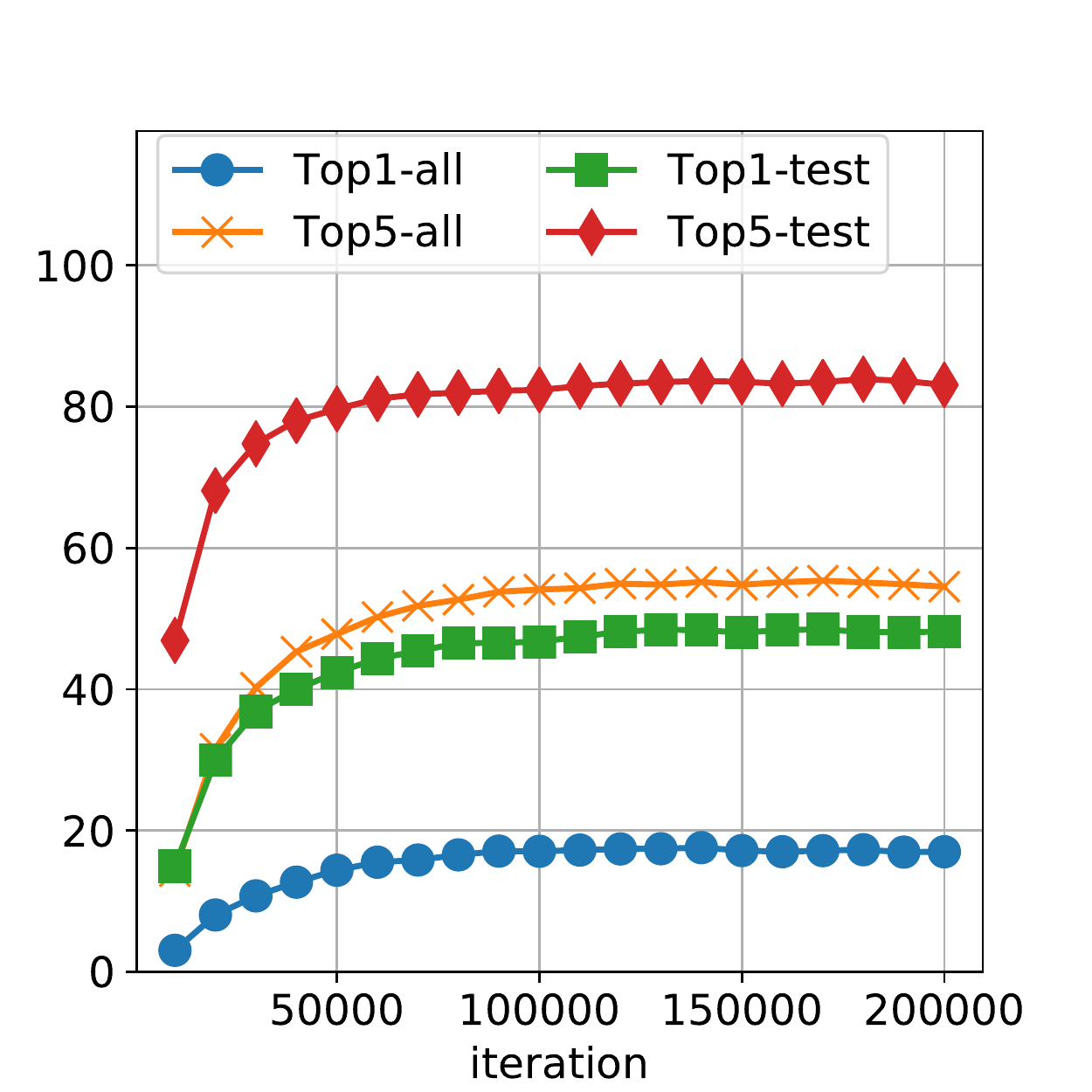}
	\includegraphics[trim=0.00in 0.0in 0.3in 0in,width=0.24\textwidth]{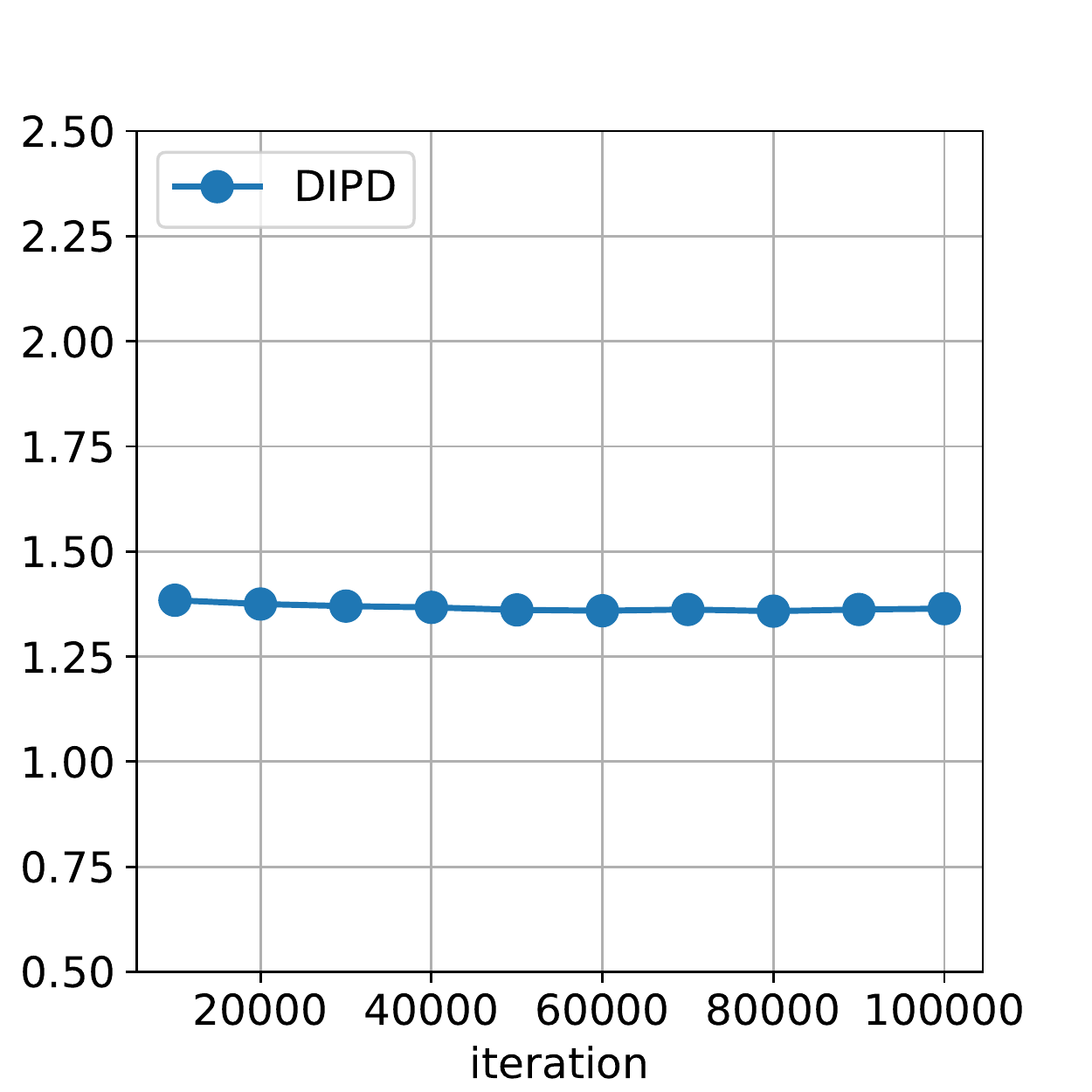}
	\includegraphics[trim=0.00in 0.0in 0.3in 0in,width=0.24\textwidth]{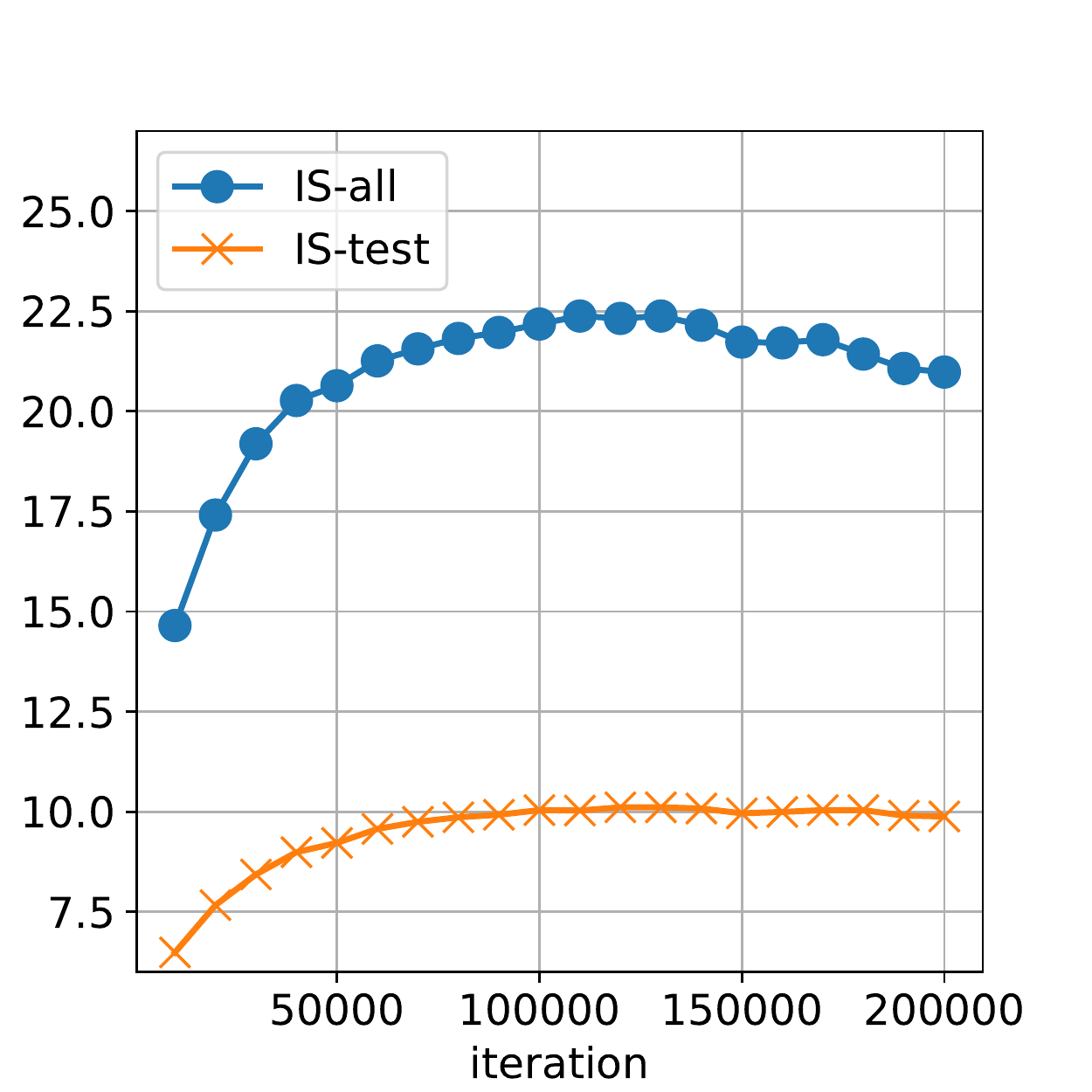}
	\includegraphics[trim=0.00in 0.0in 0.3in 0in,width=0.24\textwidth]{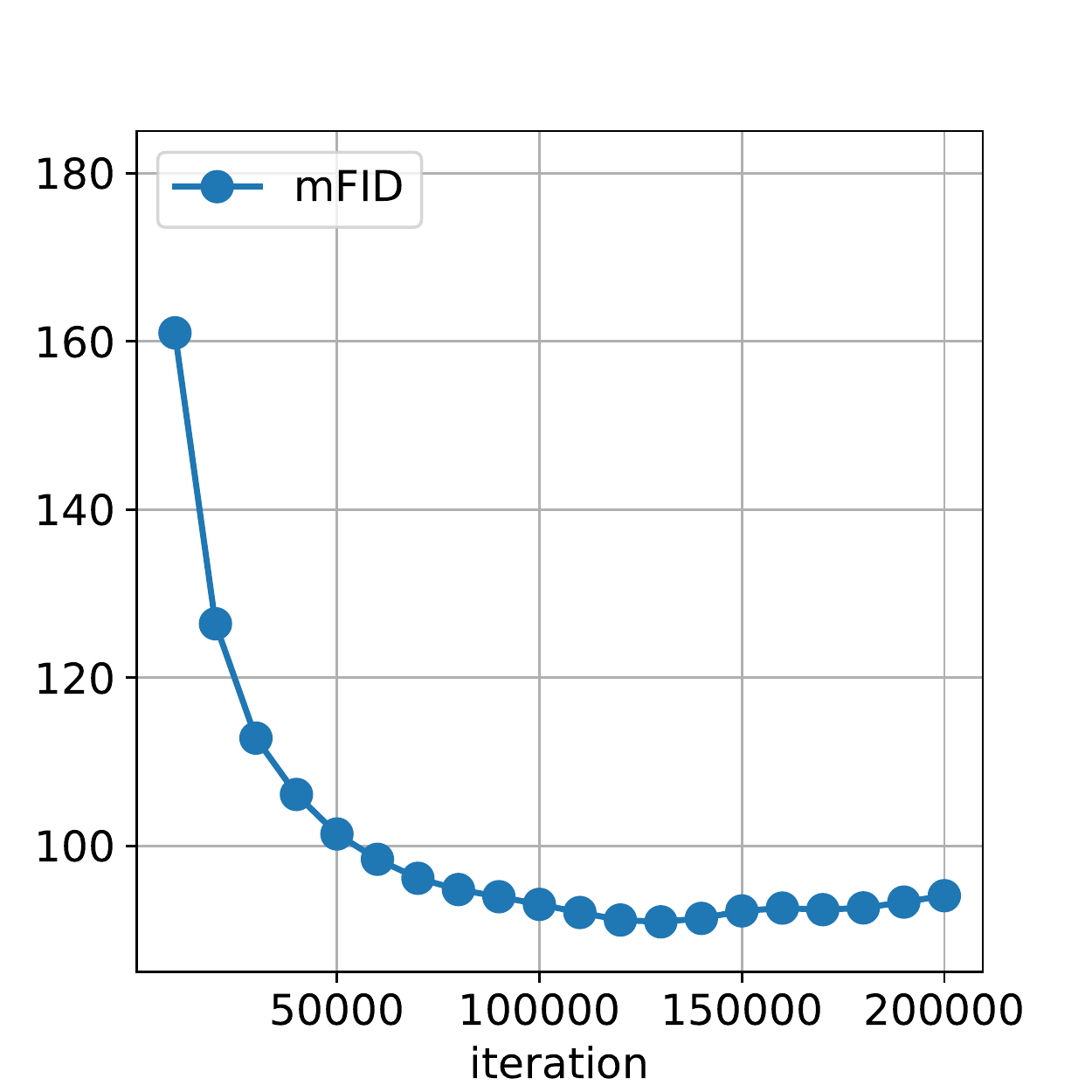}
	\centering
	\includegraphics[trim=0.00in 0.0in 0.3in 0in,width=0.24\textwidth]{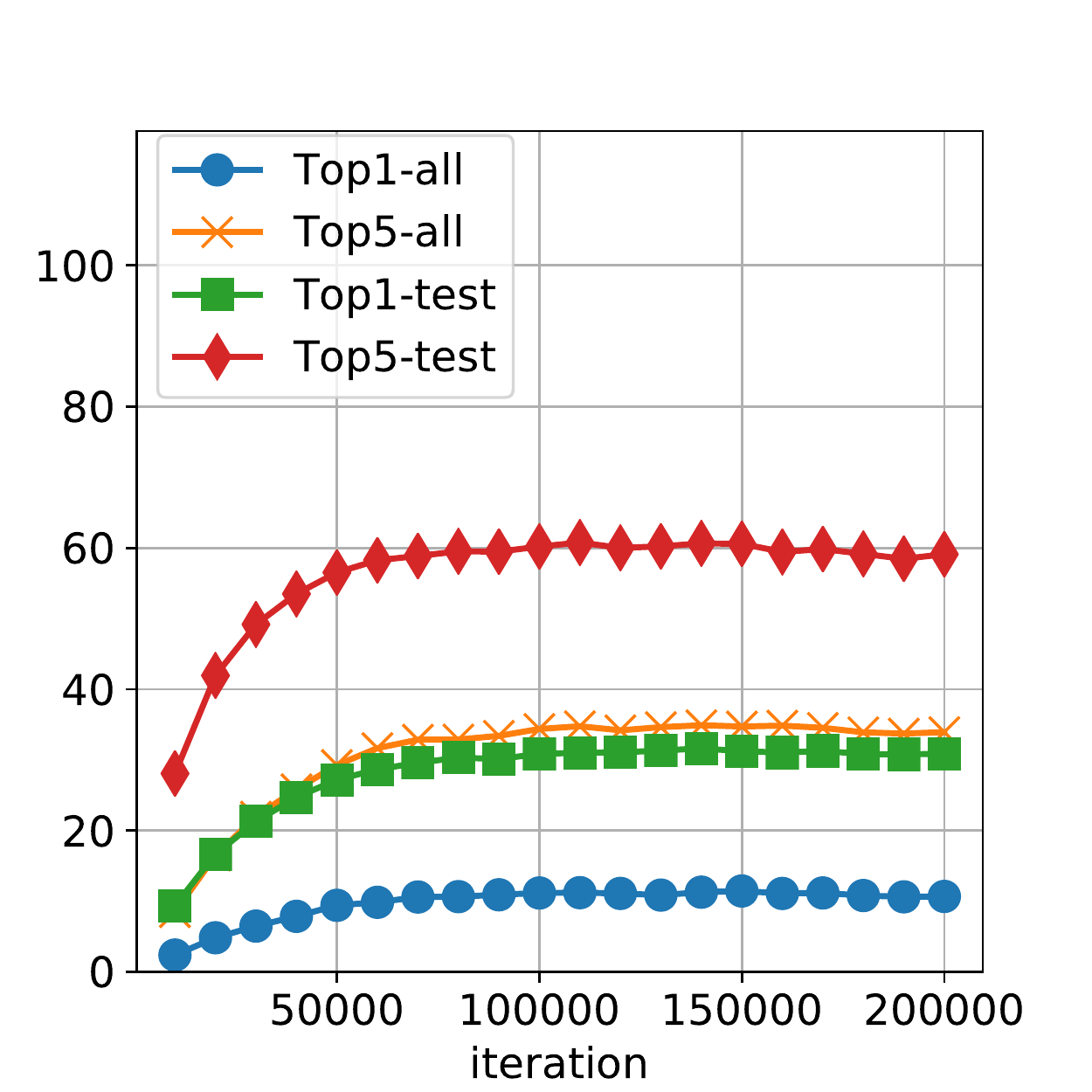}
	\includegraphics[trim=0.00in 0.0in 0.3in 0in,width=0.24\textwidth]{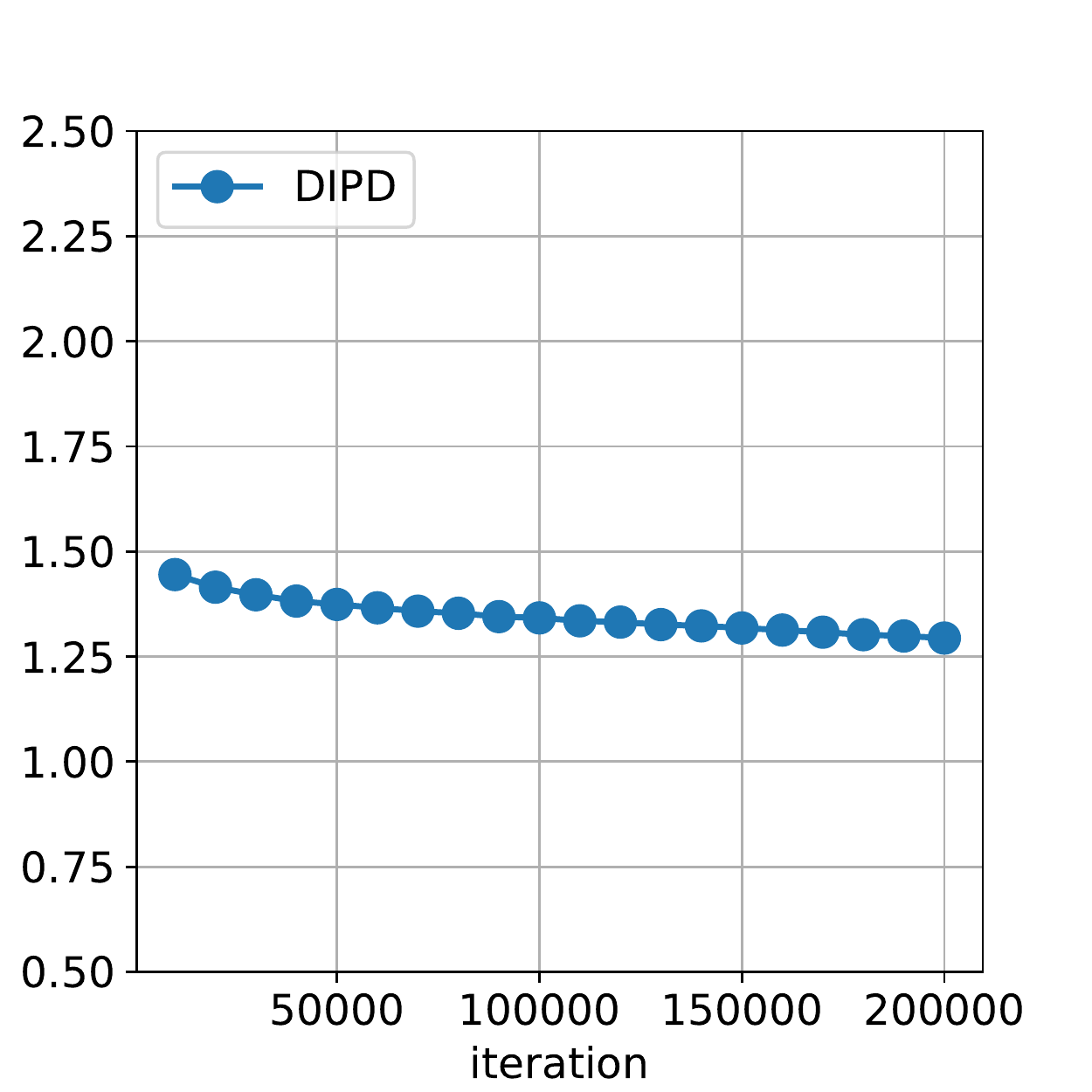}
	\includegraphics[trim=0.00in 0.0in 0.3in 0in,width=0.24\textwidth]{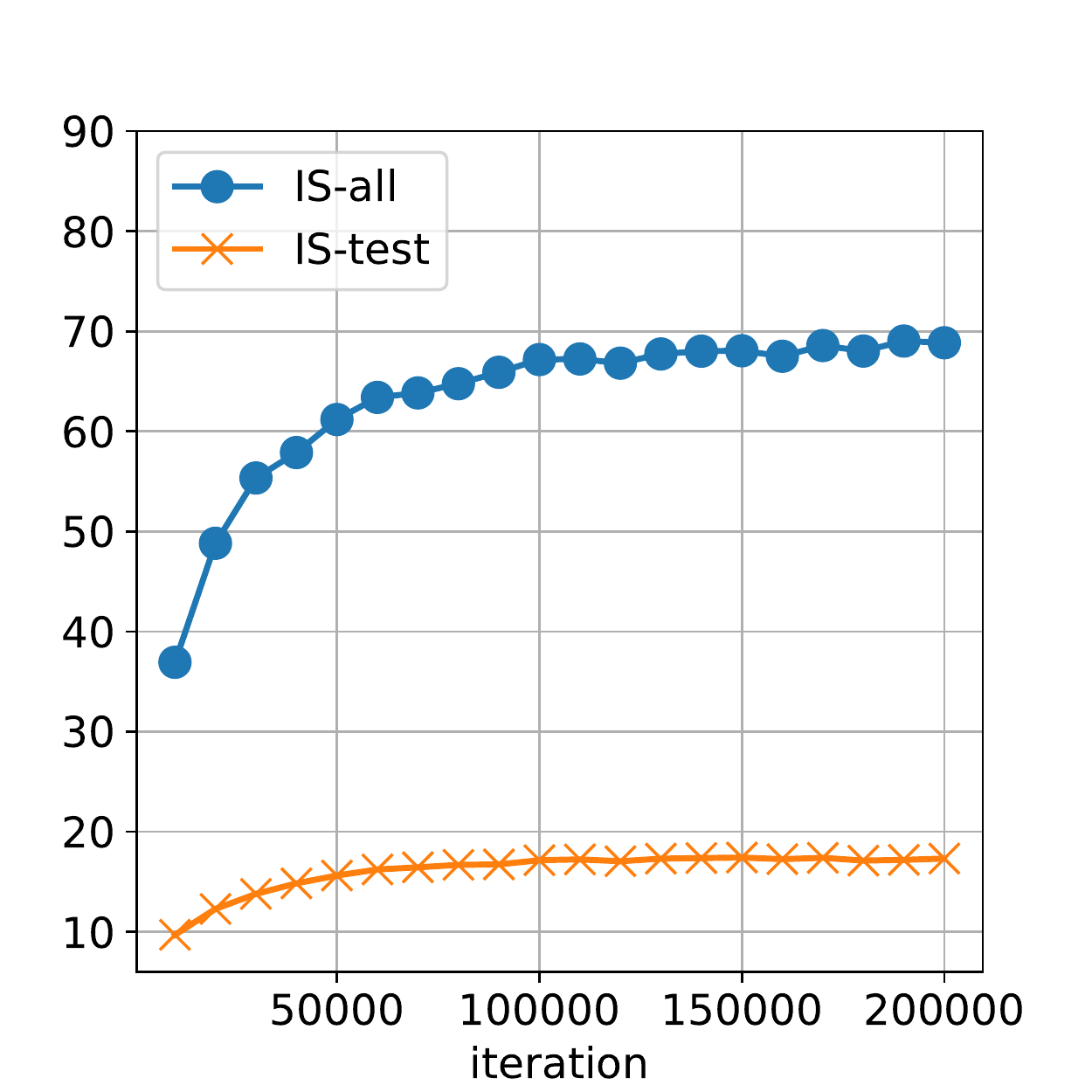}
	\includegraphics[trim=0.00in 0.0in 0.3in 0in,width=0.24\textwidth]{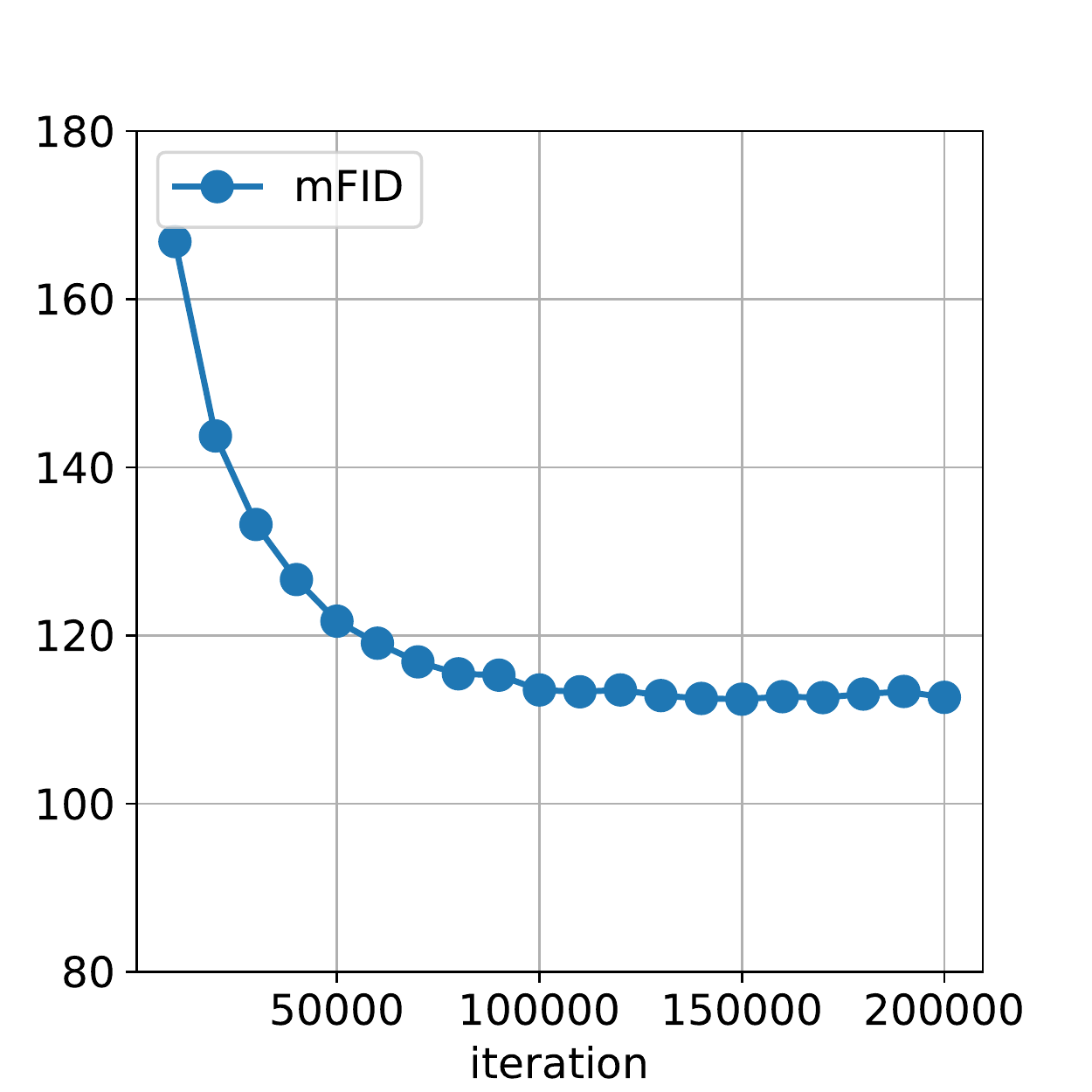}
	\scaption{Few-shot image translation performance vs. training iterations. Top row: results on the Animal Faces dataset; bottom row: results on the North American Birds dataset.}
	\label{fig::iteration}\vspace{-2mm}
\end{figure*}
\begin{figure}[t!]	
	\centering
	\includegraphics[trim=0.00in 0.0in 0.0in 0.0in,width=\columnwidth]{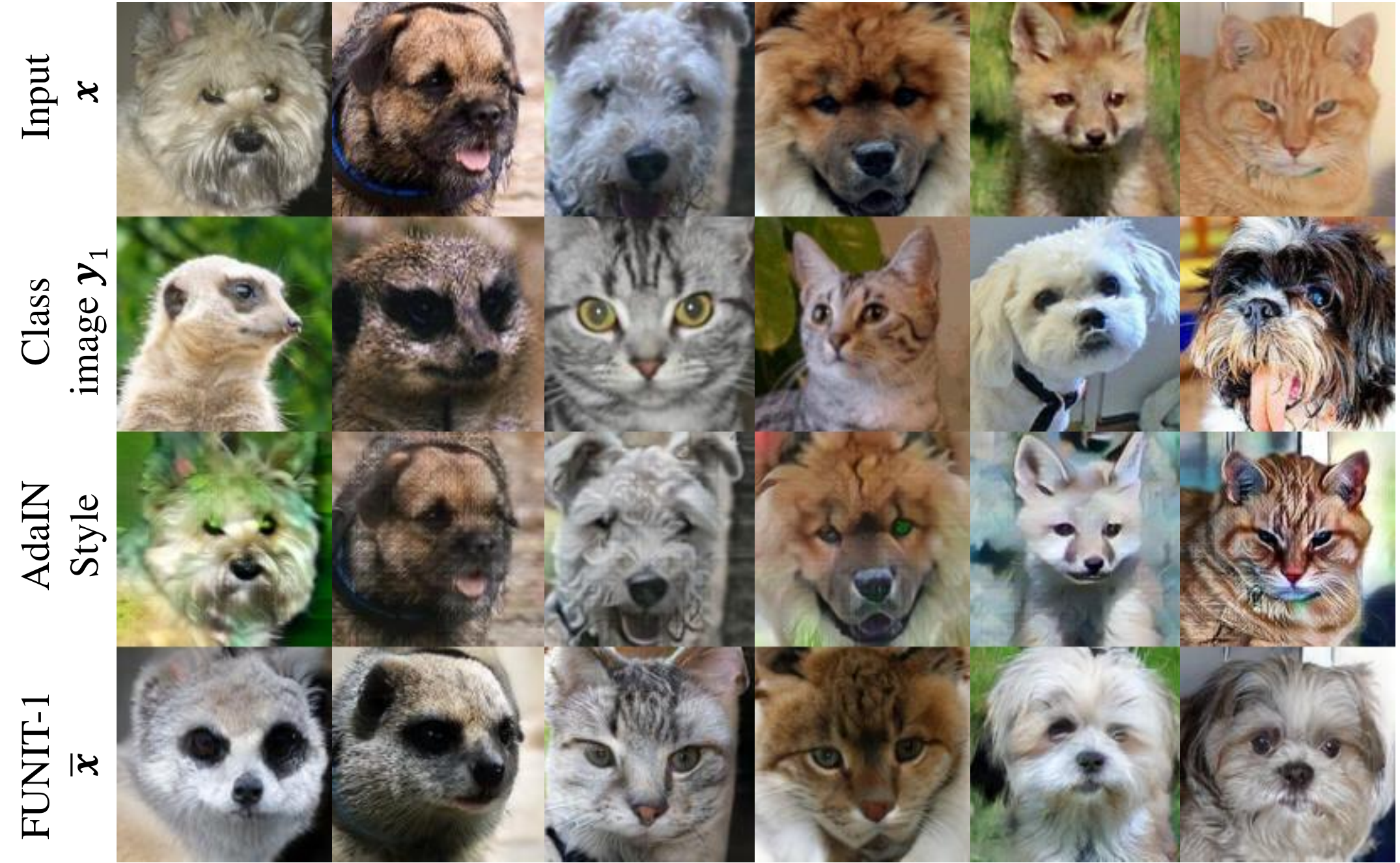}
	\scaption{\texttt{FUNIT-1} versus AdaIN style transfer~\cite{huang2017adain} for few-shot image translation.}
	\label{fig::vis_adain}
\end{figure}
\begin{figure}[t!]
	\centering
	\includegraphics[trim=0.00in 0.0in 0.0in 0in,width=0.49\textwidth]{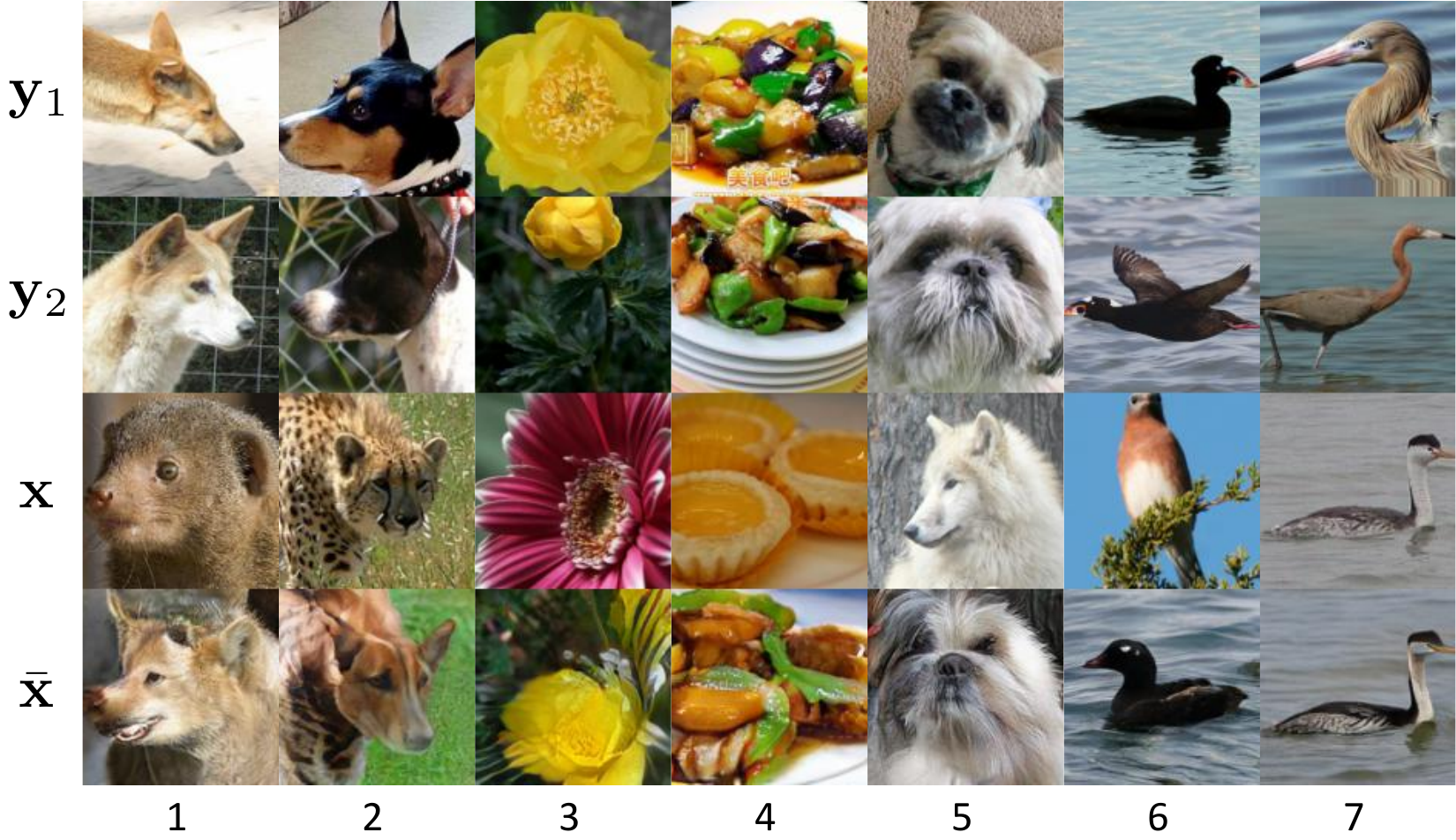}
	\scaption{Failure cases. The typical failure cases of the proposed \proposed model include generating hybrid objects (\eg column 1, 2, 3, and 4), ignoring input content images (\eg column 5 and 6), and ignoring input class images (\eg column 7).}
	\label{fig::vis_failure}\vspace{-5mm}
\end{figure}

\begin{figure*}[t!]	
	\centering
	\includegraphics[trim=0in 0in 0in 0in,width=0.99\textwidth]{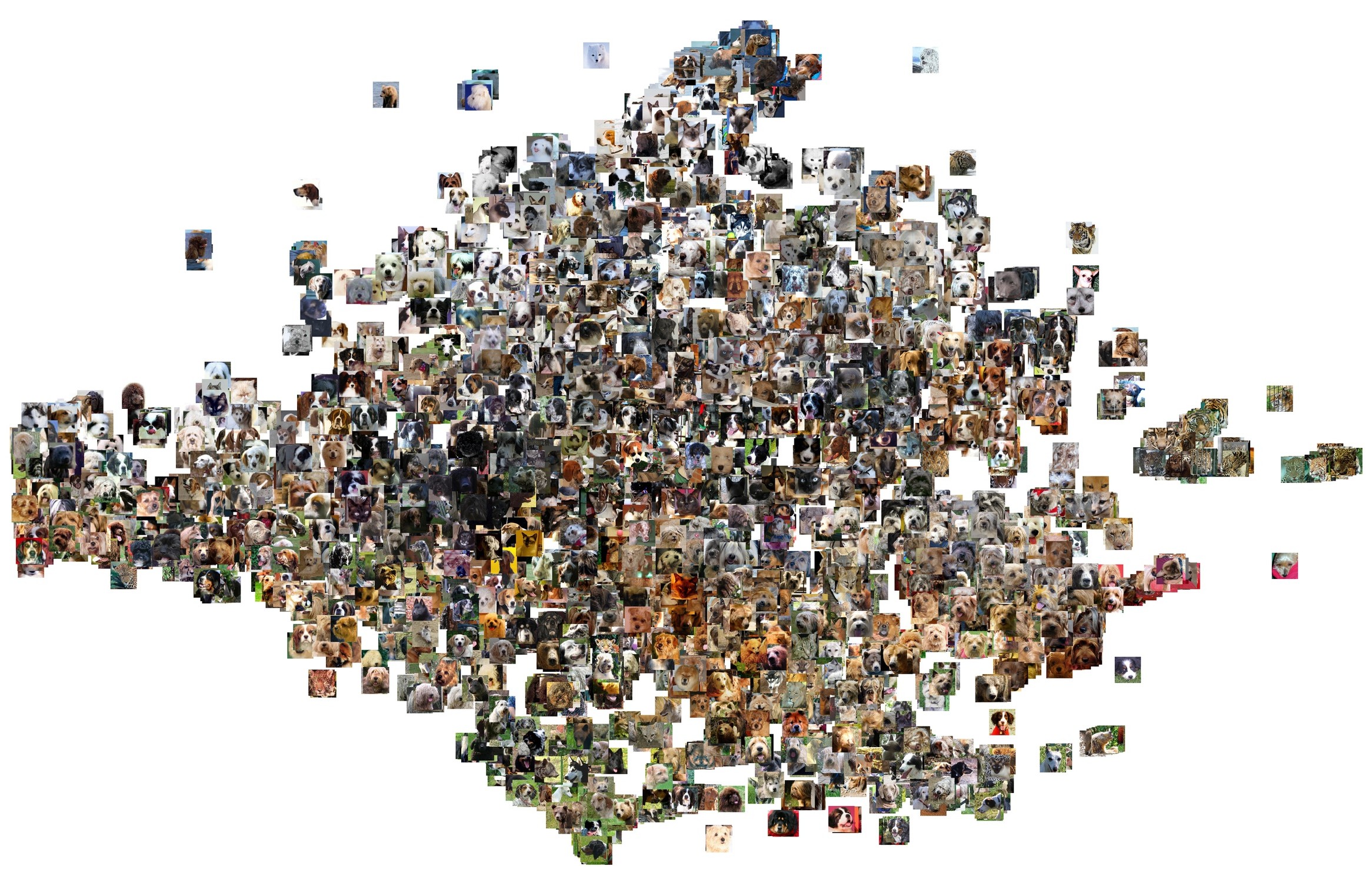}
	\scaption{2-D representation of the class code using t-SNE for 5000 images across 50 source classes. Please zoom-in for details.}
	\label{fig::supp_tsne}
	\vspace{2mm}
	\centering
	\includegraphics[trim=0in 0in 0in 0in,width=0.85\textwidth]{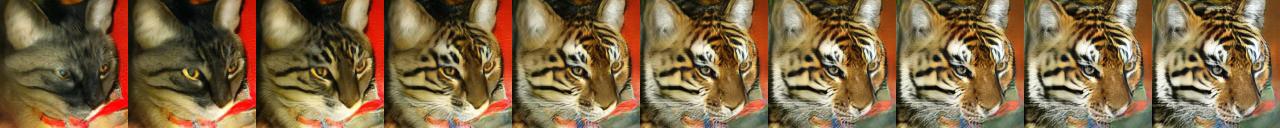}
	\includegraphics[trim=0in 0in 0in 0in,width=0.85\textwidth]{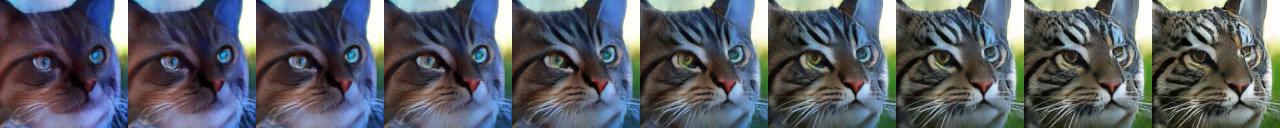}
	\includegraphics[trim=0in 0in 0in 0in,width=0.85\textwidth]{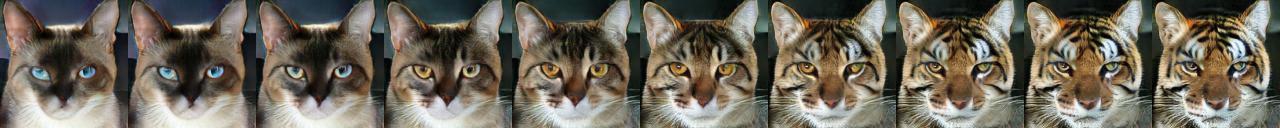}
	\includegraphics[trim=0in 0in 0in 0in,width=0.85\textwidth]{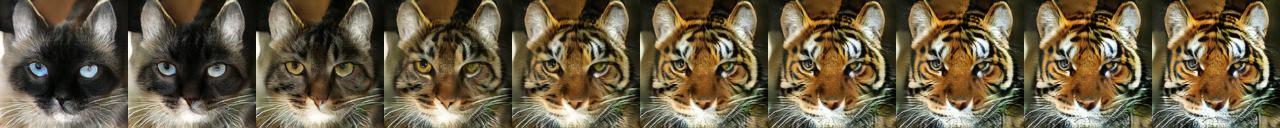}
	\includegraphics[trim=0in 0in 0in 0in,width=0.85\textwidth]{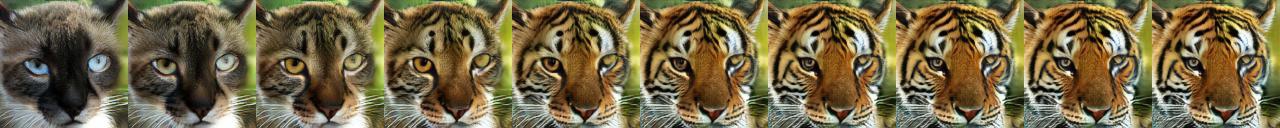}
	\includegraphics[trim=0in 0in 0in 0in,width=0.85\textwidth]{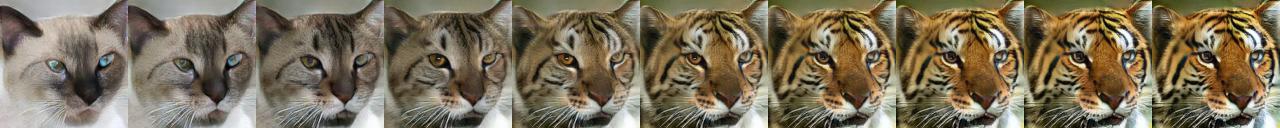}
	\scaption{Interpolation by keeping the content code fixed while interpolating between two class codes of source classes. }
	\label{fig::supp_interp}
\end{figure*}

In the main paper, we show that the few-shot translation performance is positively correlated with number of source classes in the training set for the animal face translation task. In Figure~\ref{fig::object_class_nabirds}, we show this is the same case for the bird translation task. Specifically, we report performance of the proposed model versus number of available source classes in the training set using the North American Birds dataset. We vary the number of source classes from 189, 222, 289, 333, 389, to 444 and plot the performance scores. We find the curves of the scores follow the same trend as those of the Animal Faces dataset shown in the main paper. When the model sees a larger number of source classes during training, it performs better during testing.

\section{Ablation Study}\label{sec::ablation}

\begin{table*}[t!]
	\centering
	\small
	\begin{tabular}{|c||c|c|c|c|c|c|c|}
		\hline 
		\multirow{2}{*}{\textbf{Method}} & \textbf{\# of generated} & \multicolumn{5}{c|}{\textbf{Split}} & \textbf{Average} \\
		& \textbf{Samples} & \multicolumn{1}{c}{\textbf{1}} & \multicolumn{1}{c}{\textbf{2}} & \multicolumn{1}{c}{\textbf{3}} & \multicolumn{1}{c}{\textbf{4}} & \multicolumn{1}{c|}{\textbf{5}} & \textbf{Accuracy} \\
		\hline
		\textbf{Baseline} & 0 & 38.81 $\pm$ 0.01 & 41.99 $\pm$ 0.03 & 39.13 $\pm$ 0.01 & 37.05 $\pm$ 0.02 & 36.82 $\pm$ 0.01 & 38.76 \\
		\hline
		\multirow{3}{*}{\proposed} & 10 & 41.20 $\pm$ 0.41 & 46.25 $\pm$ 0.27 & 42.65 $\pm$ 0.41 & 40.75 $\pm$ 0.20 & 39.39 $\pm$ 0.31 & 42.05 \\
		& 50 & 41.24 $\pm$ 0.16 & 46.27 $\pm$ 0.07 & 43.15 $\pm$ 0.06 & 41.01 $\pm$ 0.19 & 39.43 $\pm$ 0.09 & 42.22 \\
		& 100 & 41.01 $\pm$ 0.18 & 46.72 $\pm$ 0.05 & 42.89 $\pm$ 0.09 & 40.73 $\pm$ 0.20 & 39.33 $\pm$ 0.04 & 42.14 \\
		\hline
		\multirow{3}{*}{\texttt{S\&H}~\cite{hariharan2017low}} & 10 & 39.87 $\pm$ 0.47 & 42.69 $\pm$ 0.34 & 41.42 $\pm$ 0.39 & 39.95 $\pm$ 0.58 & 38.64 $\pm$ 0.42 & 40.51 \\& 50 & 39.93 $\pm$ 0.15 & 42.62 $\pm$ 0.28 & 40.89 $\pm$ 0.09 & 39.31 $\pm$ 0.17 & 38.44 $\pm$ 0.13 & 40.24 \\
		& 100 & 40.05 $\pm$ 0.31 & 41.72 $\pm$ 0.19 & 41.29 $\pm$ 0.16 & 41.33 $\pm$ 0.21 & 39.39 $\pm$ 0.16 & 40.76 \\
		\hline
	\end{tabular}\vspace{1mm}
	\scaption{One-shot accuracies on the 5 splits of the Animal Faces dataset when using generated images and 1 real image. The average accuracy over 5 independent runs is reported per split (different set of generated images is sampled each time).}
	\label{tbl::animals_all_splits}	
	\vspace{3mm}
	\centering
	\small
	\begin{tabular}{|c||c|c|c|c|c|c|c|}
		\hline 
		\multirow{2}{*}{\textbf{Method}} & \textbf{\# of generated} & \multicolumn{5}{c|}{\textbf{Split}} & \textbf{Average} \\
		& \textbf{Samples} & \multicolumn{1}{c}{\textbf{1}} & \multicolumn{1}{c}{\textbf{2}} & \multicolumn{1}{c}{\textbf{3}} & \multicolumn{1}{c}{\textbf{4}} & \multicolumn{1}{c|}{\textbf{5}} & \textbf{Accuracy} \\
		\hline
		\textbf{Baseline} & 0 & 30.71 $\pm$ 0.02 & 29.04 $\pm$ 0.01 & 31.93 $\pm$ 0.01 & 29.59 $\pm$ 0.01 & 30.64 $\pm$ 0.02 & 30.38 \\
		\hline
		\multirow{3}{*}{\proposed} & 10 & 32.94 $\pm$ 0.49 & 33.29 $\pm$ 0.25 & 35.15 $\pm$ 0.22 & 31.20 $\pm$ 0.20 & 34.48 $\pm$ 0.58 & 33.41 \\
		& 50 & 32.92 $\pm$ 0.34 & 33.78 $\pm$ 0.25 & 35.04 $\pm$ 0.10 & 31.80 $\pm$ 0.14 & 34.66 $\pm$ 0.17 & 33.64 \\
		& 100 & 33.83 $\pm$ 0.16 & 33.99 $\pm$ 0.12 & 36.05 $\pm$ 0.14 & 32.01 $\pm$ 0.10 & 36.09 $\pm$ 0.19 & 34.39 \\
		\hline
		\multirow{3}{*}{\texttt{S\&H}~\cite{hariharan2017low}} & 10 & 30.55 $\pm$ 0.11 & 31.96 $\pm$ 0.30 & 34.18 $\pm$ 0.19 & 30.65 $\pm$ 0.09 & 31.49 $\pm$ 0.24 & 31.77 \\
		& 50 & 31.39 $\pm$ 0.07 & 30.59 $\pm$ 0.11 & 33.60 $\pm$ 0.05 & 30.92 $\pm$ 0.20 & 31.81 $\pm$ 0.18 & 31.66 \\
		& 100 & 30.83 $\pm$ 0.10 & 32.03 $\pm$ 0.09 & 34.39 $\pm$ 0.17 & 31.12 $\pm$ 0.10 & 32.23 $\pm$ 0.15 & 32.12 \\
		\hline
	\end{tabular}\vspace{1mm}
	\scaption{One-shot accuracies on the 5 splits of the North Amercian Birds dataset when using generated images and 1 real image. The average accuracy over 5 independent runs is reported per split (different set of generated images is sampled each time).}
	\label{tbl::nabirds_all_splits}		
\end{table*}
\begin{table}[t!]
	\centering
	\small
    \resizebox{\columnwidth}{!}{
	\begin{tabular}{|c||c|c|c|c|c|c|c|c|c|c|c|c|c|c|}
		\hline 
		\multirow{3}{*}{\textbf{Method}} & \textbf{\# of} & \multicolumn{2}{c|}{\textbf{Average}}\\
		& \textbf{generated} & \multicolumn{2}{c|}{\textbf{Accuracy}}\\
		\cline{3-4}
		& {\bf samples} & {\bf Animal Faces} &{\bf N. A. Birds} \\
		\hline
		Prototypical Net.~[39] & 0 & 34.51 & 28.49 \\
		\hline
		Prototypical Net. & 10 & 40.07 & 33.63 \\
		$+$ & 50 & 39.85 & 33.74 \\
		{\proposed}& 100 & 40.47 & 34.28 \\
		\hline
	\end{tabular}%
	}\vspace{2mm}
	\scaption{1-shot accuracies on the Animal Faces and the North American Birds datasets upon using generated images with the prototypical networks method~[39], averaged over 5 splits.}
	\label{tbl::prototypical}
\end{table}

In Table~\ref{tbl::lambda_r}, we analyze impact of the content image reconstruction loss weight on the Animal Faces dataset. We find that a larger $\lambda_R$ value leads to a smaller domain-invariant perceptual distance with the expense of a lower translation accuracy. The table shows that $\lambda_R=0.1$ provides a good trade-off, and we used it as the default value throughout the paper. Interestingly, a very small weight value $\lambda_R=0.01$ results in degrading performance on both content preservation and translation accuracy. This indicates that the content reconstruction loss help regularize the training.

Table~\ref{tbl::ablation} presents results of an ablation study analyzing impact of the loss terms in the proposed algorithm on the Animal Faces dataset. We find that removing the feature matching loss term resulting in a slightly degraded performance. But when removing the zero-centered gradient penalty, both content preservation and translation accuracy degrade a lot.

In Figure~\ref{fig::iteration}, we plot performance of the proposed model over training iterations on the one-shot setting (\texttt{FUNIT-1}). The translation accuracy, content preservation, image quality, and distribution matching scores improve with more iterations in general. The improvement is more dramatic in the early stage and slows down around 10000 iterations. We hence use 10000 iterations as the default parameter for reporting experiment results throughout the paper.

\section{Comparison with AdaIN Style Transfer}\label{sec::adain}

In Figure~\ref{fig::vis_adain}, we compare the proposed method with the AdaIN style transfer method~\cite{huang2017adain} for the few-shot animal face translation task. While the AdaIN style transfer method can change the textures of the input animals, it does not change their shapes. As a result, the translation outputs still resemble to the inputs in terms of appearances.

\section{Failure Case}\label{sec::failure}

Figure~\ref{fig::vis_failure} illustrates several failure cases of the proposed algorithm. They include generating hybrid classes, ignoring input content images, and ignoring input class images.

\section{Latent Space Interpolation}\label{sec::latent}

\begin{table*}[t!]
	\resizebox{1.00\linewidth}{!}{\mbox{
\begin{tabular}{|c|c||cccc|c|cc|c|}
	\hline
	& Setting & \bf Top1-all $\uparrow$ & \bf Top5-all $\uparrow$ & \bf Top1-test $\uparrow$ & \bf Top5-test $\uparrow$ & \bf DIPD $\downarrow$& \bf IS-all $\uparrow$ & \bf IS-test $\uparrow$ & \bf mFID $\downarrow$\\
	\hline
	\parbox[t]{2mm}{\multirow{5}{*}{\rotatebox[origin=c]{90}{\textbf{\large{Animal}}}}}
	& \texttt{CycleGAN-Unfair-All}&  46.02&  71.51&  63.44&  89.60&  1.417&  19.91&  12.77&  90.34\\
	& \texttt{UNIT-Unfair-All}&  41.93&  67.19&  60.76&  88.42&  {\bf1.388}&  20.18&  12.29&  92.22\\
	& \texttt{MUNIT-Unfair-All}&  {66.25}&  {86.59}&  {78.45}&  95.10&  1.670&  21.99&  {16.68}&  71.18\\
	& \texttt{StarGAN-Unfair-All}&  39.72&  66.18&  61.24&  89.52&  1.392&  13.29&  9.79&  141.57 \\
	& \texttt{FUNIT-Unfair-All}&  {\bf 83.00}&  {\bf 89.12}&  {\bf 91.08}&  {\bf 99.88}&  1.445&  \textbf{25.92}&  {\bf 23.45}&  {\bf 32.73} \\ 		
	\hline
\end{tabular}}}
\vspace{1mm}
\scaption{Comparison to the competing methods when using all the images for training.}\label{tbl::vs_all}
\end{table*}

We explore the latent space learned by the class encoder. In Figure~\ref{fig::supp_tsne}, we use t-SNE to visualize the class code in a two dimensional space. It can be seen that images from similar classes are grouped together in the class embedding space. 

Figure~\ref{fig::supp_interp} shows interpolation results by keeping the content code fixed while interpolating the class code between those of two source class images. Interestingly, we find that by interpolating between two source classes (Siamese cat and Tiger) we can sometimes generate a target class (Tabby cat) that the model has never observed. This suggests that the class encoder learns a general class-specific representation, thus enabling generalization to novel classes.

\section{Few-Shot Classification}\label{sec::classification}

As mentioned in the main paper, we conduct an experiment using images generated by the \proposed generator to train classifiers for novel classes in the one-shot setting, using the Animal Faces and North American Birds datasets. Following the setup in Hariharan~\etal~\cite{hariharan2017low}, we create 5 different one-shot training splits where each has a training, validation, and test set. The training set consists of $|\mathbb{T}|$ images, one image from each of the $|\mathbb{T}|$ test classes. The validation set consists of 20-100 images from each test classes. The test set consists of remaining test class images. 

We use the \proposed generator to generate a synthetic training set by using the images in the classification training set as the class image input and randomly sampled images from the source classes as the content image input. We train a classifier using both of the original and synthetic training sets. We compare our method against the Shrink and Hallucinate (\texttt{S\&H}) method of Hariharan~\etal~\cite{hariharan2017low}, which learns to generate final layer features corresponding to novel classes. We use a pretrained 10-layer ResNet network as the feature extractor, which is pretrained purely using the source class images, and train a linear classifier over target classes. We find it crucial to weight the loss on generated images lower than that on real images. We conduct an exhaustive grid search on the weight value as well as the weight decay value using the validation set and report the performance on the test set. For a fair comparison, we also perform the same exhaustive search for the S\&H method. 

In Table~\ref{tbl::few_shot_classification} of the main paper, we report performance of our method and the \texttt{S\&H} method~\cite{hariharan2017low} over different number of generated samples (\ie, images for \proposed and features for the \texttt{S\&H}) on two challenging fine-grained classification tasks. Both methods perform better than the baseline classifier that uses just the single provided real image per novel class. Using our generated images, we obtain around 2\% improvement over the \texttt{S\&H} method that generates features.

The base 10-layer ResNet network is trained for 90 epochs, with an initial learning rate of 0.1 decayed by a factor of 10 every 30 epochs. Weight decay for the linear classifier over novel classes is chosen from 15 logarithmically spaced values between and including 0.000001 and 0.1. The loss multiplier for loss on generated images and features is chosen from 7 logarithmically spaced values between and including 0.001 and 1. The values for weight decay and loss multiplier are chosen based on the best validation set accuracy obtained while training on Split \#1. These values are then fixed and used for all remaining splits 2-5. The task of learning an L2 regularized classifier using fixed features is a convex optimization problem, and we use line search with the L-BFGS algorithm, and thus do not have to specify a learning rate.

In Tables~\ref{tbl::animals_all_splits} and~\ref{tbl::nabirds_all_splits}, we report test accuracies and their associated variances on one-shot learning for all 5 one-shot splits of the Animal Faces and the North American Birds datasets. In all experiments, we only learn a new classifier layer using features extracted from a network trained on the set of classes used to train the image generator.

Our method can also be used in conjunction with existing few-shot classification approaches. In Table~\ref{tbl::prototypical}, we show 1-shot classification results obtained using the Prototypical Networks method~\cite{snell2017prototypical}, which assigns to a test sample the label of the closest prototype (cluster center) obtained from the given few train samples. Clearly, using our generated samples together with the 1 provided sample per class at test time to compute class prototype representations helps improve accuracy on both datasets considered by over 5.5\%.

\section{More Translation Results}\label{sec::more}

\begin{figure*}[!t]
	\centering
	\includegraphics[trim=0.00in 0.0in 0in 0in,width=0.99\textwidth] {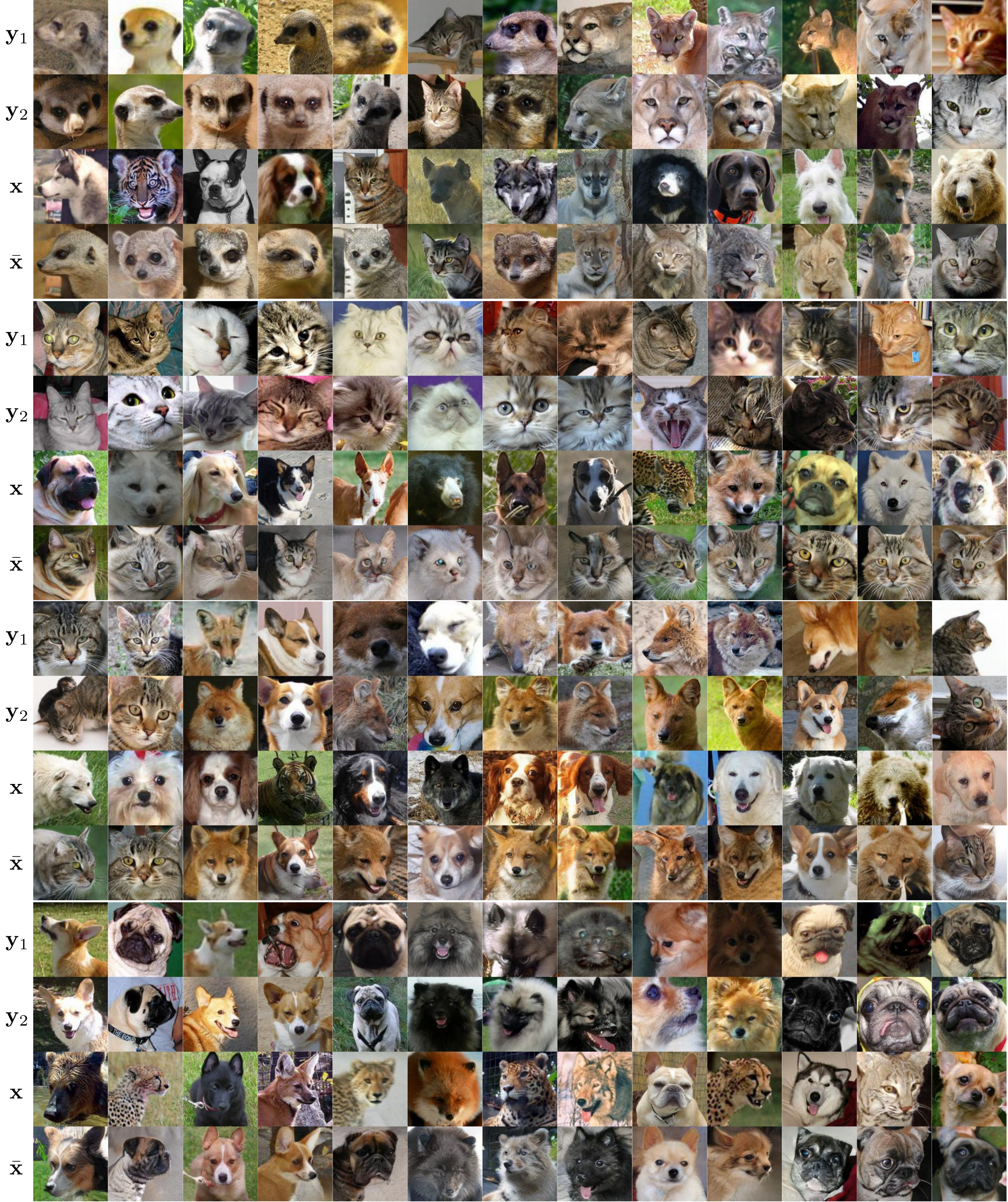}
	\scaption{Additional visualization results on the few-shot animal face image translation task. All the results are computed using the same \texttt{FUNIT-5} model. The model can be re-purposed for generating images of a dynamically specified target class in the test time by having access to 5 images from the target class. The variable $\pmb{x}$ is the input content image, $\pmb{y}_1$ and $\pmb{y}_1$ are 2 out of the 5 input target class images, and $\bar{\pmb{x}}$ is the translation output. We find that the animal face in the translation output has a similar pose to the input content image but the appearance is akin to the appearance of the animal faces in the class images.} 
	\label{fig::additionl_animals}
\end{figure*}

\begin{figure*}[!t]
	\centering
	\includegraphics[trim=0.00in 0.0in 0in 0in,width=0.99\textwidth] {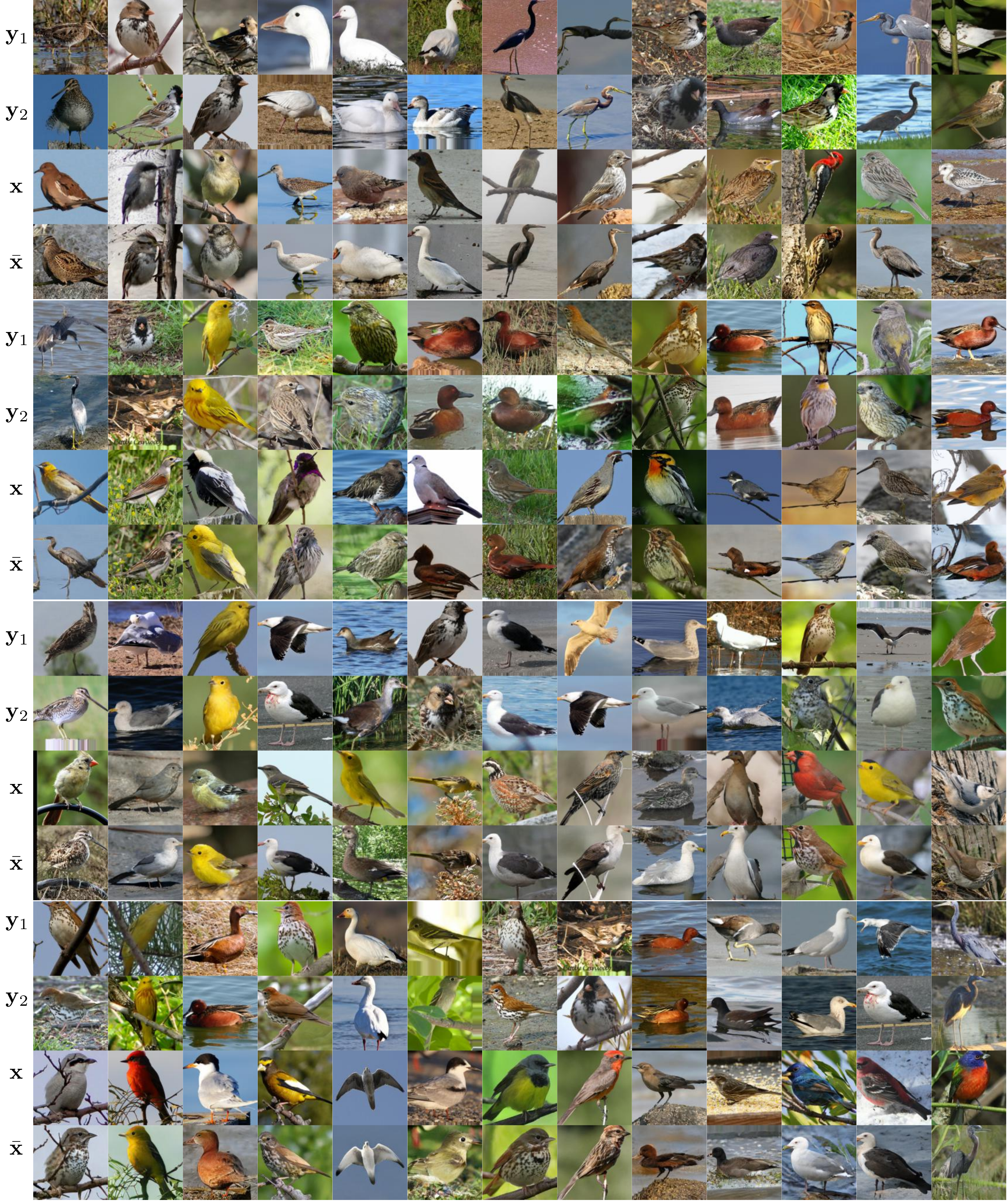}
	\scaption{Additional visualization results on the few-shot bird image translation task. All the results are computed using the same \texttt{FUNIT-5} model. The model can be re-purposed for generating images of a dynamically specified target class in the test time by having access to 5 images from the target class. The variable $\pmb{x}$ is the input content image, $\pmb{y}_1$ and $\pmb{y}_1$ are 2 out of the 5 input target class images, and $\bar{\pmb{x}}$ is the translation output. We find that the bird in the translation output has a similar pose to the input content image but the appearance is akin to the appearance of the birds in the class images.} 
	\label{fig::additionl_birds}
\end{figure*}

\begin{figure*}[!t]
	\centering
	\includegraphics[trim=0.00in 0.0in 0in 0in,width=0.99\textwidth] {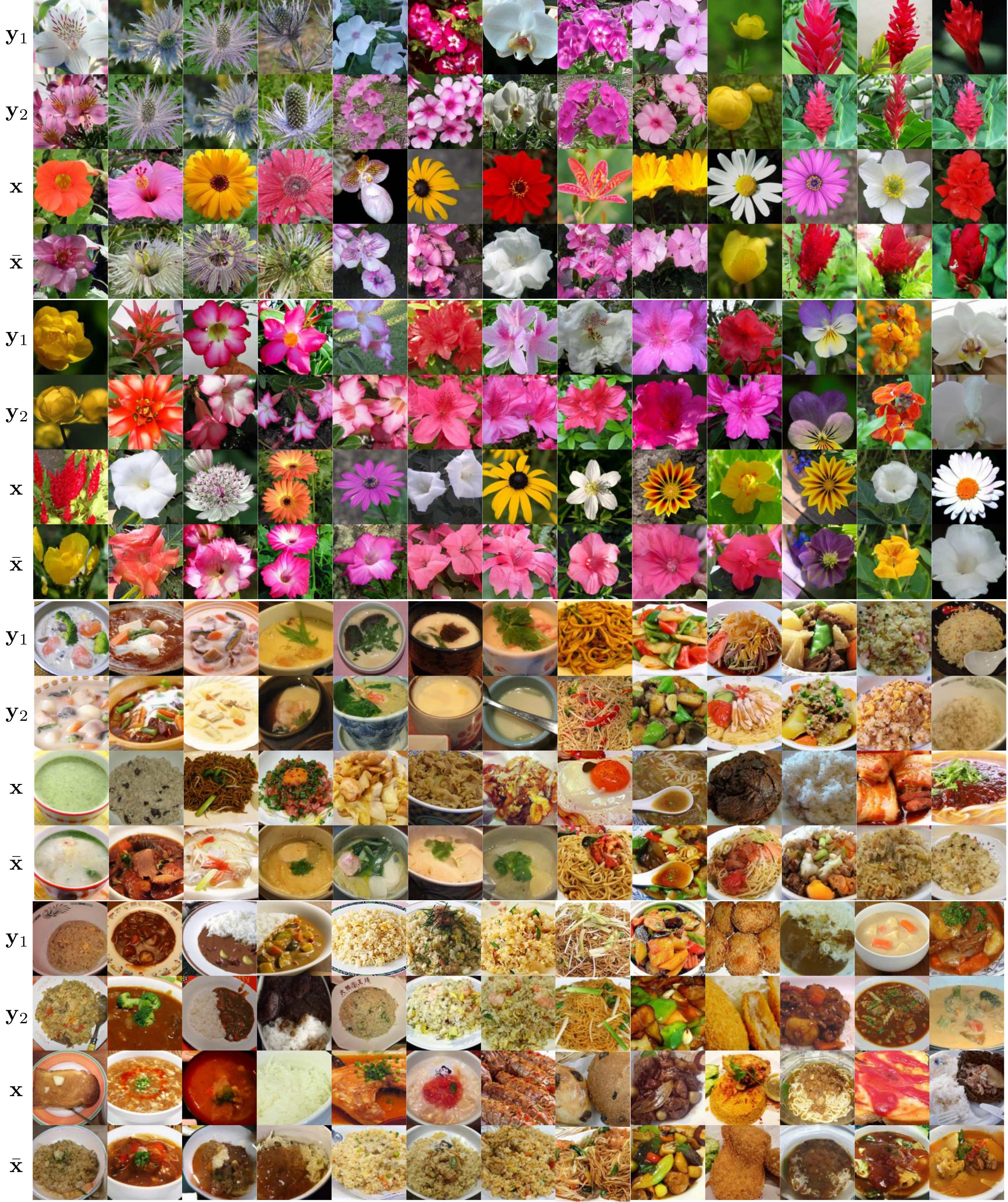}
	\scaption{Additional visualization results on the few-shot flower and food image translation tasks. All the results for the same task are computed using the same \texttt{FUNIT-5} model. The model can be re-purposed for generating images of a dynamically specified target class in the test time by having access to 5 images from the target class. The variable $\pmb{x}$ is the input content image, $\pmb{y}_1$ and $\pmb{y}_1$ are 2 out of the 5 input target class images, and $\bar{\pmb{x}}$ is the translation output. For flower translation, we find the flowers in the output and input image have a similar pose. For food translation, the bowl and plate remain at the same location while the food are changed from one kind to the other.} 
	\label{fig::additionl_flowers_and_foods}
\end{figure*}

\begin{figure*}[!t]
	\centering
	\includegraphics[trim=0.00in 0.0in 0in 0in,width=0.99\textwidth] {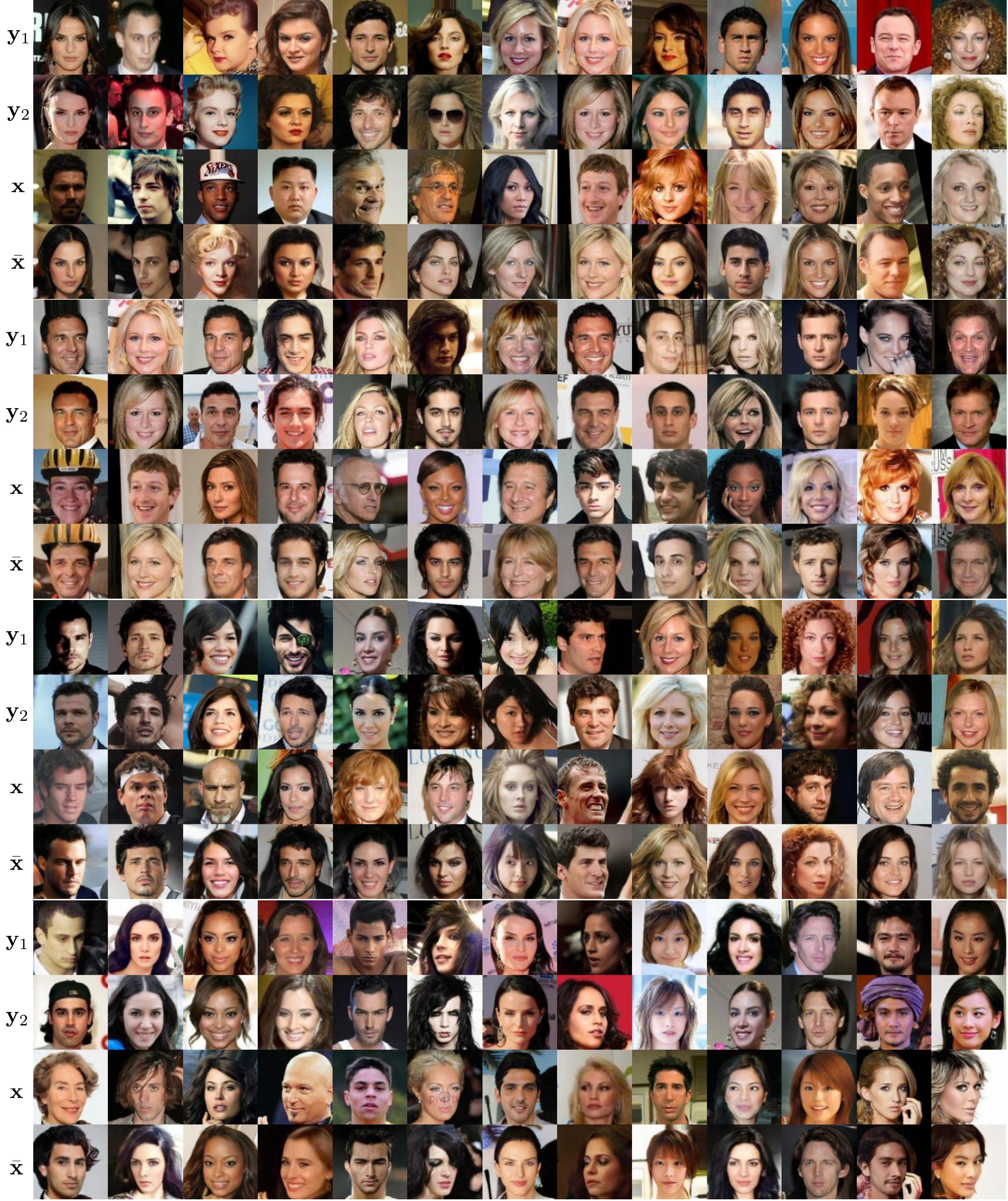}
	\scaption{Visualization results on the few-shot face image translation task. All the results are computed using the same \texttt{FUNIT-5} model. The model can be re-purposed for generating images of a dynamically specified target class in the test time by having access to 5 images from the target class. The variable $\pmb{x}$ is the input content image, $\pmb{y}_1$ and $\pmb{y}_1$ are 2 out of the 5 input target class images, and $\bar{\pmb{x}}$ is the translation output. We find that the face in the translation output has a similar pose to the input content image but the appearance is akin to the appearance of the faces of the target person.} 
	\label{fig::additionl_faces}
\end{figure*}

In Figure~\ref{fig::additionl_animals},~\ref{fig::additionl_birds},~\ref{fig::additionl_flowers_and_foods}, and~\ref{fig::additionl_faces}, we show additional few-shot translation results for the animal face image translation task, the bird image translation task, the flower image translation task, the food image translation task, and the face image translation task. For the face translation, each class is defined by a person identify. The experiment is conducted using the celebrity face dataset~\cite{liu2015faceattributes}. All the results are computed using \texttt{FUNIT-5}.

\section{FUNIT-All}\label{sec::funit_all}

To evaluate performance of the proposed method in the standard image-to-image translation setting where target classes images are available in the training time, we trained a \proposed model using all the training images in the animal face dataset. The resulting model is called \texttt{FUNIT-Unfair-All}. In this setting, there is no unseen classes in the test time. As shown in Table~\ref{tbl::vs_all}, our \texttt{FUNIT-Unfair-All} outperforms state-of-the-art image-to-image translation models including CycleGAN~\cite{zhu2017unpaired}, UNIT~\cite{liu2017unsupervised}, MUNIT~\cite{huang2018multimodal}, and StarGAN~\cite{choi2017stargan}. (For models that can only handle translation between two domains. We trained multiple of them for evaluation.) This shows that the proposed \proposed model is also a competitive multi-class image-to-image translation model.
\end{document}